\documentclass{article}

\PassOptionsToPackage{numbers, compress}{natbib}

\usepackage[final]{neurips_2022}

\usepackage[utf8]{inputenc} 
\usepackage[T1]{fontenc}    
\usepackage{hyperref}       
\usepackage{url}            
\usepackage{booktabs}       
\usepackage{amsfonts}       
\usepackage{nicefrac}       
\usepackage{microtype}      
\usepackage[table]{xcolor}         
\usepackage{multirow}
\usepackage{graphicx}
\usepackage{amsmath}
\usepackage{makecell}

\usepackage{enumitem}
\usepackage{cancel}
\usepackage{listings}

\usepackage{xcolor}
\definecolor{codegreen}{rgb}{0,0.6,0}
\definecolor{codegray}{rgb}{0.5,0.5,0.5}
\definecolor{codepurple}{rgb}{0.58,0,0.82}
\definecolor{backcolour}{rgb}{0.95,0.95,0.92}

\lstdefinestyle{mystyle}{
    backgroundcolor=\color{backcolour},   
    commentstyle=\color{codegreen},
    keywordstyle=\color{magenta},
    numberstyle=\tiny\color{codegray},
    stringstyle=\color{codepurple},
    basicstyle=\ttfamily\footnotesize,
    breakatwhitespace=false,         
    breaklines=true,                 
    captionpos=b,                    
    keepspaces=true,                 
    numbers=left,                    
    numbersep=5pt,                  
    showspaces=false,                
    showstringspaces=false,
    showtabs=false,                  
    tabsize=2
}
\usepackage{pifont}
\newcommand{\cmark}{\text{\ding{51}}}%
\newcommand{\xmark}{\text{\ding{55}}}%
\usepackage{amsthm}
\newtheorem{prop}{Proposition}

\newtheorem{thm}{Theorem}

\title{RankFeat: Rank-1 Feature Removal for Out-of-distribution Detection}

\author{%
  Yue Song, Nicu Sebe, Wei Wang \\
  Department of Information Engineering and Computer Science \\
  University of Trento, Italy \\ 
  \texttt{yue.song@unitn.it} \\
}

\begin{document}

\maketitle

\begin{abstract}
The task of out-of-distribution (OOD) detection is crucial for deploying machine learning models in real-world settings. In this paper, we observe that the singular value distributions of the in-distribution (ID) and OOD features are quite different: the OOD feature matrix tends to have a larger dominant singular value than the ID feature, and the class predictions of OOD samples are largely determined by it. This observation motivates us to propose \texttt{RankFeat}, a simple yet effective \texttt{post hoc} approach for OOD detection by removing the rank-1 matrix composed of the largest singular value and the associated singular vectors from the high-level feature (\emph{i.e.,} $\mathbf{X}{-} \mathbf{s}_{1}\mathbf{u}_{1}\mathbf{v}_{1}^{T}$). \texttt{RankFeat} achieves the \emph{state-of-the-art} performance and reduces the average false positive rate (FPR95) by 17.90\% compared with the previous best method. Extensive ablation studies and comprehensive theoretical analyses are presented to support the empirical results.
\end{abstract}

\section{Introduction}


In the real-world applications of deep learning, understanding whether a test sample belongs to the same distribution of training data is critical to the safe deployment of machine learning models. The main challenge stems from the fact that current deep learning models can easily give over-confident predictions for out-of-distribution (OOD) data~\cite{nguyen2015deep}. Recently a rich line of literature has emerged to address the challenge of OOD detection~\cite{wang2021can,huang2021importance,bibas2021single,diffenderfer2021winning,sun2021react,gibbs2021adaptive,fort2021exploring,sutter2021robust,ye2021towards,kumar2022fine,garg2022leveraging,zaeemzadeh2021out,du2022vos,gomes2022igeood,haroush2021statistical}.

\begin{figure}[t]
    \centering
    \includegraphics[width=0.9\linewidth]{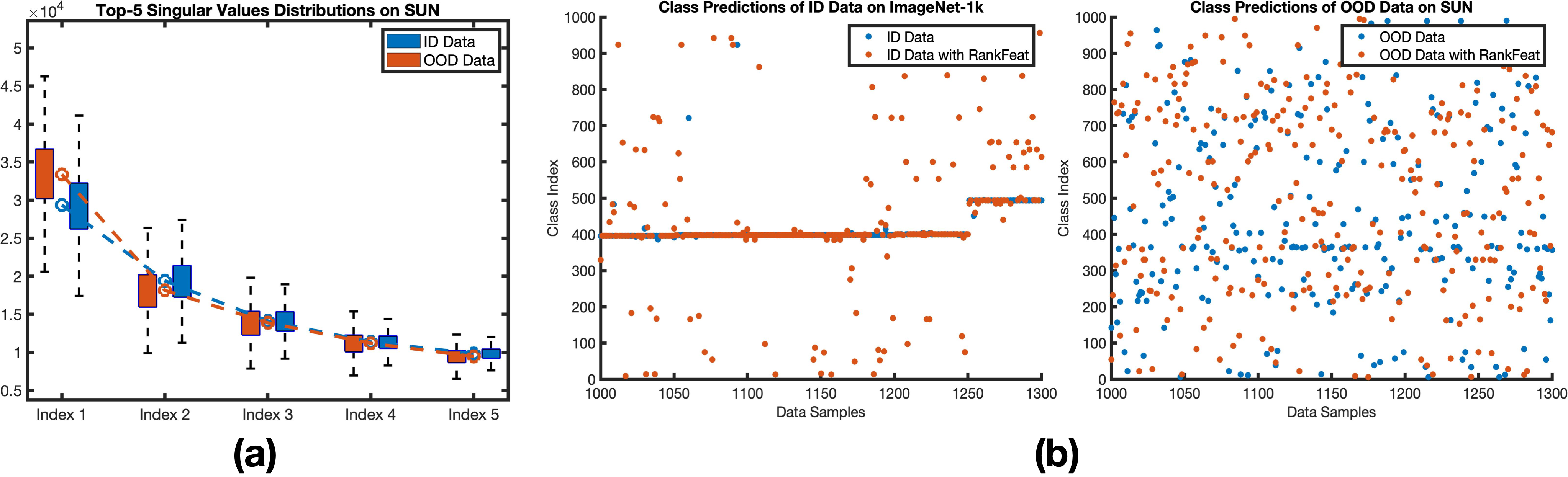}
    \caption{\textbf{(a)} The distribution of top-5 singular values for the ID and OOD features on ImageNet-1k and SUN. The OOD feature matrix tends to have a significantly larger dominant singular value. \textbf{(b)} After removing the rank-1 matrix composed by the dominant singular value and singular vectors, the class predictions of OOD data are severely perturbed, while those of ID data are moderately influenced. This observation indicates that the decisions of OOD data heavily depend on the dominant singular value and the corresponding singular vectors of the feature matrix. In light of this finding, we get motivated to propose \texttt{RankFeat} for OOD detection by removing the rank-1 matrix from the high-level feature. Both observations also hold for other OOD datasets.}
    \label{fig:cover}
\end{figure}

Previous OOD detection approaches either rely on the feature distance~\cite{lee2018simple}, activation abnormality~\cite{sun2021react}, or gradient norm~\cite{huang2021importance}. In this paper, we tackle the problem of OOD detection from another perspective: by analyzing the spectrum of the high-level feature matrices (\emph{e.g.,} the output of Block 3 or Block 4 of a typical ResNet~\cite{he2016deep} model), we observe that the feature matrices have quite different singular value distributions for the in-distribution (ID) and OOD data (see Fig.~\ref{fig:cover}(a)): \textit{the OOD feature tends to have a much larger dominant singular value than the ID feature, whereas the magnitudes of the rest singular values are very similar.} This peculiar behavior motivates us to remove the rank-1 matrix composed of the dominant singular value and singular vectors from the feature. As displayed in Fig.~\ref{fig:cover}(b), removing the rank-1 feature drastically perturbs the class prediction of OOD samples; a majority of predictions have been changed. On the contrary, most ID samples have consistent classification results before and after removing the subspace. \textit{This phenomenon indicates that the over-confident prediction of OOD samples might be largely determined by the dominant singular value and the corresponding singular vectors.}

\begin{figure}[t]
    \centering
    \includegraphics[width=0.99\linewidth]{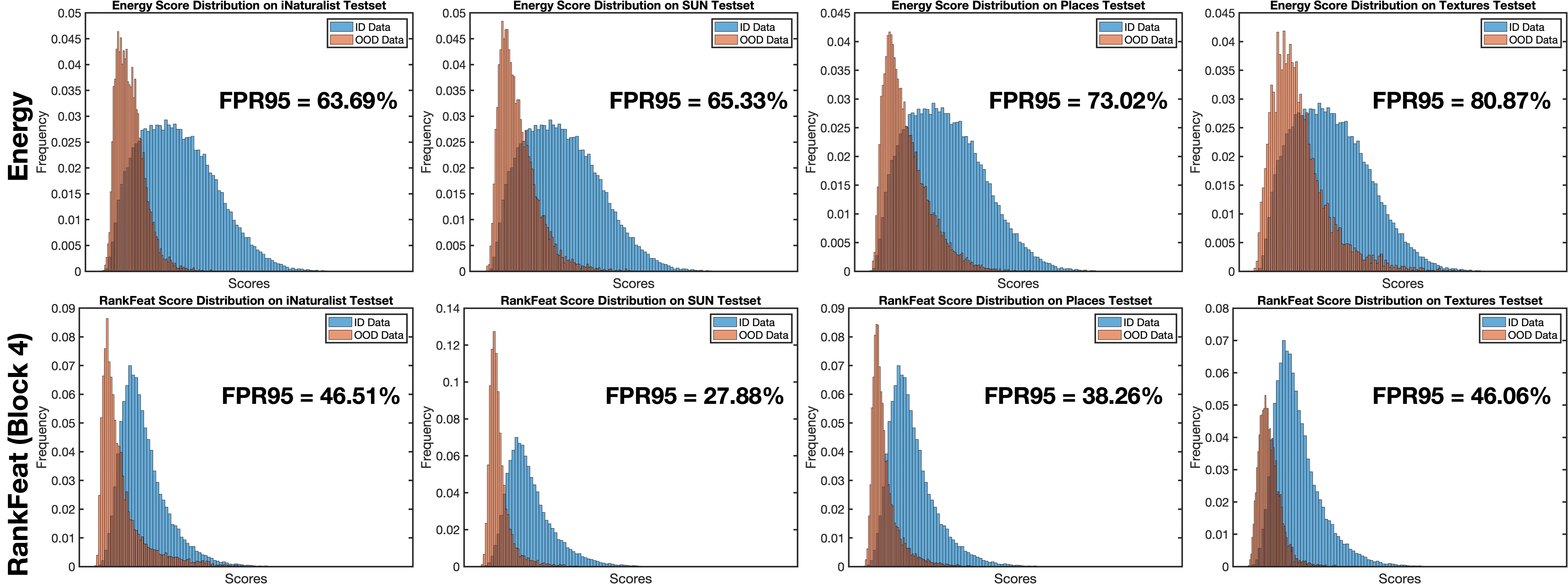}
    \caption{The score distributions of \texttt{Energy}~\cite{liu2020energy} (top row) and our proposed \texttt{RankFeat} (bottom row) on four OOD datasets. Our method can better separate the ID and OOD data. }
    \label{fig:score_dist}
\end{figure}

Based on this observation, we \textbf{assume} that the first singular value of OOD feature tends to be much larger than that of ID feature. The intuition behind is that the OOD feature corresponds to a larger PCA explained variance ratio (being less informative), and the well-trained network weights might cause and amplify the difference (see Sec. E of the supplementary for the detailed illustration). Hence, we conjecture that leveraging this gap might help to better distinguish ID and OOD samples. To this end, we propose \texttt{RankFeat}, a simple but effective \textit{post hoc} approach for OOD detection. \texttt{RankFeat} perturbs the high-level feature by removing its rank-1 matrix composed of the dominant singular value and vectors. Then the logits derived from the perturbed features are used to compute the OOD score function. By removing the rank-1 feature, the over-confidence of OOD samples is mitigated, and consequently the ID and OOD data can be better distinguished (see Fig.~\ref{fig:score_dist}). Our \texttt{RankFeat} establishes the \textit{state-of-the-art} performance on the large-scale ImageNet benchmark and a suite of widely used OOD datasets across different network depths and architectures. In particular, \texttt{RankFeat} outperforms the previous best method by \textbf{17.90\%} in the average false positive rate (FPR95) and by \textbf{5.44\%} in the area under curve (AUROC). Extensive ablation studies are performed to reveal important insights of \texttt{RankFeat}, and comprehensive theoretical analyses are conducted to explain the working mechanism. Code is publicly available via \url{https://github.com/KingJamesSong/RankFeat}.

The \textbf{key results and main contributions} are threefold:
\begin{itemize}
    \item We propose \texttt{RankFeat}, a simple yet effective \textit{post hoc} approach for OOD detection by removing the rank-1 matrix from the high-level feature. \texttt{RankFeat} achieves the \textit{state-of-the-art} performance across benchmarks and models, reducing the average FPR95 by \textbf{17.90\%} and improving the average AUROC by \textbf{5.44\%} compared to the previous best method. 
    \item We perform extensive ablation studies to illustrate the impact of (1) removing or keeping the rank-1 matrix, (2) removing the rank-n matrix (n>1), (3) applying our \texttt{RankFeat} at various network depths, (4) the number of iterations to iteratively derive the approximate rank-1 matrix for acceleration but without performance degradation, and (5) different fusion strategies to combine multi-scale features for further performance improvements.
    \item Comprehensive theoretical analyses are conducted to explain the working mechanism and to underpin the superior empirical results. We show that (1) removing the rank-1 matrix reduces the upper bound of OOD score more, (2) removing the rank-1 matrix makes the statistics of OOD feature closer to random matrices, and (3) both \texttt{RankFeat} and \texttt{ReAct}~\cite{sun2021react} work by optimizing the upper bound containing the largest singular value. \texttt{ReAct}~\cite{sun2021react} indirectly and manually clips the underlying term, while \texttt{RankFeat} directly subtracts it.
\end{itemize}


\section{RankFeat: Rank-1 Feature Removal for OOD Detection}

In this section, we introduce the background of OOD detection task and our proposed \texttt{RankFeat} that performs the OOD detection by removing the rank-1 matrix from the high-level feature.

\noindent \textbf{Preliminary: OOD detection.} The OOD detection is often formulated as a binary classification problem with the goal to distinguish between ID and OOD data. Let $f$ denote a model trained on samples from the ID data $\mathcal{D}_{in}$. For the unseen OOD data $\mathcal{D}_{out}$ at test time, OOD detection aims to define a decision function $\mathcal{G}(\cdot)$:
\begin{equation}
    \mathcal{G}(\mathbf{x})=
    \begin{cases}
    {\rm in} & \mathcal{S}(\mathbf{x})>\gamma ,\\
    {\rm out} & \mathcal{S}(\mathbf{x})<\gamma .
    \end{cases}
\end{equation}
where $\mathbf{x}$ denotes the data encountered at the inference stage, $\mathcal{S}(\cdot)$ is the seeking scoring function, and $\gamma$ is a chosen threshold to make a large portion of ID data correctly classified (\emph{e.g.,} $95\%$). The difficulty of OOD detection lies in designing an appropriate scoring function $\mathcal{S}(\cdot)$ such that the score distributions of ID and OOD data overlap as little as possible. 

\noindent \textbf{RankFeat: rank-1 feature removal.} Consider the reshaped high-level feature map $\mathbf{X}{\in}\mathbb{R}^{C{\times}HW}$ of a deep network (the batch size is omitted for simplicity). Here 'high-level feature' denotes the feature map that carries rich semantics in the later layers of a network (\emph{e.g.,} the output of Block 3 or Block 4 of a typical deep model like ResNet). 
Our \texttt{RankFeat} first performs the Singular Value Decomposition (SVD) on each individual feature matrix in the mini-batch to decompose the feature:
\begin{equation}
    \mathbf{X} = \mathbf{U}\mathbf{S}\mathbf{V}^{T}
\end{equation}
where $\mathbf{S}{\in}\mathbb{R}^{C{\times}HW}$ is the rectangular diagonal singular value matrix, and $\mathbf{U}{\in}\mathbb{R}^{C{\times}C}$ and $\mathbf{V}{\in}\mathbb{R}^{HW{\times}HW}$ are left and right orthogonal singular vector matrices, respectively. Then \texttt{RankFeat} removes the rank-1 matrix from the feature as:
\begin{equation}
    \mathbf{X}' = \mathbf{X} - \mathbf{s}_{1}\mathbf{u}_{1}\mathbf{v}_{1}^{T}
\end{equation}
where $\mathbf{s}_{1}$ is the largest singular value, and $\mathbf{u}_{1}$ and $\mathbf{v}_{1}$ are the corresponding left and right singular vectors, respectively. The perturbed feature is fed into the rest of the network to generate the logit predictions $\mathbf{y}'$. Finally, \texttt{RankFeat} computes the energy score of the logits for the input $\mathbf{x}$ as:
\begin{equation}
    {\texttt{RankFeat}}(\mathbf{x})=\log\sum\exp(\mathbf{y}')
\end{equation}
By removing the rank-1 matrix composed by the dominant singular value $\mathbf{s}_{1}$, the over-confident predictions of OOD data are largely perturbed. In contrast, the decisions of ID data are mildly influenced. This could help to separate the ID and OOD data better in the logit space.




\noindent \textbf{Acceleration by Power Iteration.} Since \texttt{RankFeat} only involves the dominant singular value and vectors, there is no need to compute the full SVD of the feature matrix. Hence our method can be potentially accelerated by Power Iteration (PI). The PI algorithm is originally used to approximate the dominant eigenvector of a Hermitian matrix. With a slight modification, it can also be applied to general rectangular matrices. Given the feature $\mathbf{X}$, the modified PI takes the coupled iterative update:
\begin{equation}
    \mathbf{v}_{k}=\frac{\mathbf{X}\mathbf{u}_{k}}{||\mathbf{X}\mathbf{u}_{k}||},\  \mathbf{u}_{k+1}=\Big(\frac{\mathbf{v}_{k}^{T}\mathbf{X}}{||\mathbf{v}_{k}^{T}\mathbf{X}||}\Big)^{T}
\end{equation}
where $\mathbf{u}_{0}$ and $\mathbf{v}_{0}$ are initialized with random orthogonal vectors and converge
to the left and right singular vectors, respectively. After certain iterations, the dominant singular value is computed as $\mathbf{s}_{1}{=}\mathbf{v}_{k}^{T}\mathbf{X}\mathbf{u}_{k}$. As will be illustrated in Sec.~\ref{sec:exp_ablation}, the approximate solution yielded by PI achieves very competitive performance against the SVD but with much less time overhead.

\noindent \textbf{Combination of multi-scale features.} Our \texttt{RankFeat} works at various later depths of a model, \emph{i.e.,} Block 3 and Block 4. Since intermediate features might focus on different semantic information, their decision cues are very likely to be different. It is thus natural to consider fusing the scores to leverage the distinguishable information of both features for further performance improvements. Let $\mathbf{y}'$ and $\mathbf{y}''$ denote the logit predictions of Block 3 and Block 4 features, respectively. \texttt{RankFeat} performs the fusion at the logit space and computes the score function as $\log\sum\exp(\nicefrac{(\mathbf{y}'+\mathbf{y}'')}{2})$. Different fusion strategies are explored and discussed in Sec.~\ref{sec:exp_ablation}.
\section{Theoretical Analysis}
\label{sec:theory}

In this section, we perform some theoretical analyses on \texttt{RankFeat} to support the empirical results. We start by proving that removing the rank-1 feature with a larger $\mathbf{s}_{1}$ would reduce the upper bound of \texttt{RankFeat} score more. Then based on Random Matrix Theory (RMT), we show that removing the rank-1 matrix makes the statistics of OOD features closer to random matrices. Finally, the theoretical connection of \texttt{ReAct} and our \texttt{RankFeat} is analyzed and discussed: both approaches work by optimizing the score upper bound determined by $\mathbf{s}_{1}$. \texttt{ReAct} manually uses a pre-defined threshold to clip the term with $\mathbf{s}_{1}$, whereas our \texttt{RankFeat} directly optimizes the bound by subtracting this term.


\noindent \textbf{Removing the rank-1 matrix with a larger $\mathbf{s}_{1}$ would reduce the upper bound of RankFeat more.} For our \texttt{RankFeat} score function, we can express its upper bound in an analytical form. Moreover, the upper bound analysis explicitly indicates that removing the rank-1 matrix with a larger first singular value would reduce the upper bound more. Specifically, we have the following proposition.

\begin{prop}
The upper bound of \texttt{RankFeat} score is defined as $\texttt{RankFeat}(\mathbf{x}) <\frac{1}{HW} \Big(\sum_{i=1}^{N} \mathbf{s}_{i} - \mathbf{s}_{1}\Big) ||\mathbf{W}||_{\infty} +  ||\mathbf{b}||_{\infty} + \log(Q)$ where $Q$ denotes the number of categories, and $\mathbf{W}$ and $\mathbf{b}$ are the weight and bias of the last layer, respectively. A larger $\mathbf{s}_{1}$ would reduce the upper bound more.
\end{prop}

\begin{proof}

For the feature $\mathbf{X}{\in}\mathbb{R}^{C{\times}HW}$, its SVD  $\mathbf{U}\mathbf{S}\mathbf{V}^{T}{=}\mathbf{X}$ can be expressed as the summation of rank-1 matrices $\mathbf{X}{=}\sum\mathbf{s}_{i}\mathbf{u}_{i}\mathbf{v}_{i}^{T}$. The feature perturbed by \texttt{RankFeat} can be computed as:
\begin{equation}
    \mathbf{X}' = \mathbf{X} - \mathbf{s}_{1}\mathbf{u}_{1}\mathbf{v}_{1}^{T} = \sum_{i=2}^{N} \mathbf{s}_{i}\mathbf{u}_{i}\mathbf{v}_{i}^{T}
\end{equation}
where $N$ denotes the shorter side of the matrix (usually $N{=}HW$). In most deep models~\cite{he2016deep,he2016identity}, usually the last feature map needs to pass a Global Average Pooling (GAP) layer to collapse the width and height dimensions. The GAP layer can be represented by a vector
\begin{equation}
    \mathbf{m}=\frac{1}{HW}\begin{bmatrix}1,1,\dots,1\end{bmatrix}^{T}
\end{equation}
The pooled feature map is calculated as $\mathbf{X}'\mathbf{m}$. Then the output logits are computed by the matrix-vector product with the classification head as:
\begin{equation}
\begin{gathered}
    \mathbf{y}'=\mathbf{W}\mathbf{X'}\mathbf{m}+\mathbf{b} = \sum_{i=2}^{N} (s_{i} \mathbf{W}\mathbf{u}_{i}\mathbf{v}_{i}^{T}\mathbf{m}) + \mathbf{b} 
    \label{eq:logit}
\end{gathered}
\end{equation}
where $\mathbf{W}{\in}\mathbb{R}^{W{\times}C}$ denotes the weight matrix, $\mathbf{b}{\in}\mathbb{R}^{Q{\times}1}$ represents the bias vector, and $\mathbf{y}'{\in}\mathbb{R}^{Q{\times}1}$ is the output logits that correspond to the perturbed feature $\mathbf{X}'$. Our \texttt{RankFeat} score is computed as:
\begin{equation}
    \texttt{RankFeat}(\mathbf{x}) = \log \sum_{i=1}^{Q} \exp(\mathbf{y}'_{i})
    \label{eq:rankfeat}
\end{equation}
where $\mathbf{x}$ is the input image, and $Q$ denotes the number of categories. Here we choose Energy~\cite{liu2020energy} as the base function due to its theoretical alignment with the input probability density and its strong empirical performance. Eq.~\eqref{eq:rankfeat} can be re-formulated by the \texttt{Log-Sum-Exp} trick
\begin{equation}
\begin{gathered}
     \log \sum_{i=1}^{Q} \exp(\mathbf{y}'_{i}) =  \log \sum_{i=1}^{Q} \exp(\mathbf{y}'_{i}-\max(\mathbf{y}')) + \max(\mathbf{y}')\\
\end{gathered}
\end{equation}
The above equation directly yields the tight bound as:
\begin{equation}
    \max(\mathbf{y}')   < \log \sum \exp(\mathbf{y}') < \max(\mathbf{y}') + \log(Q)
\end{equation}
Since $\max(\mathbf{y}'){\leq}\max(|\mathbf{y}'|){=}||\mathbf{y}'||_{\infty}$, we have
\begin{equation}
    \texttt{RankFeat}(\mathbf{x}) = \log \sum \exp(\mathbf{y}') < \max(\mathbf{y}') + \log(Q) \leq  ||\mathbf{y}'||_{\infty} + \log(Q) 
    \label{eq:score_upper}
\end{equation}
The vector norm has the property of triangular inequality, \emph{i.e.,} $||\mathbf{a}+\mathbf{c}||\leq||\mathbf{a}||+||\mathbf{c}||$ holds for any vectors $\mathbf{a}$ and $\mathbf{c}$. Moreover, since both $\mathbf{u}$ and $\mathbf{v}$ are orthogonal vectors, we have the relation $||\mathbf{u}_{i}||_{\infty}{\leq}1$ and $||\mathbf{v}_{i}||_{\infty}{\leq}1$.
Relying on these two properties, injecting eq.~\eqref{eq:logit} into eq.~\eqref{eq:score_upper} leads to
\begin{equation}
    \texttt{RankFeat}(\mathbf{x}) {<} \sum_{i=2}^{N} \mathbf{s}_{i} ||\mathbf{W}\mathbf{u}_{i}\mathbf{v}_{i}^{T}\mathbf{m}||_{\infty} {+} ||\mathbf{b}||_{\infty} {+} \log(Q) {\leq} \sum_{i=2}^{N} \mathbf{s}_{i} ||\mathbf{W}\mathbf{m}||_{\infty} {+}  ||\mathbf{b}||_{\infty} {+} \log(Q)
\end{equation}
Since $\mathbf{m}$ is a scaled all-ones vector, we have $||\mathbf{W}\mathbf{m}||_{\infty}{=}\nicefrac{||\mathbf{W}||_{\infty}}{HW}$. The bound is simplified as:
\begin{equation}
    \texttt{RankFeat}(\mathbf{x}) <\frac{1}{HW} \Big(\sum_{i=1}^{N} \mathbf{s}_{i} - \mathbf{s}_{1}\Big) ||\mathbf{W}||_{\infty} +  ||\mathbf{b}||_{\infty} + \log(Q)
\end{equation}
As indicated above, removing a larger $\mathbf{s}_{1}$ would reduce the upper bound of \texttt{RankFeat} score more.
\end{proof}


\noindent\textbf{Remark:} Considering that OOD feature usually has a much larger $\mathbf{s}_{1}$ (see Fig.~\ref{fig:cover}(a)), \texttt{RankFeat} would reduce the upper bound of OOD samples more. 

\emph{Notice that our bound analysis strives to improve the understanding of OOD methods from new perspectives instead of giving a strict guarantee of the score.} For example, the upper bound can be used to explain the shrinkage and skew of score distributions in Fig.~\ref{fig:score_dist}. Subtracting $\mathbf{s}_{1}$ would largely reduce the numerical range of both ID and OOD scores, which could squeeze score distributions. Since the dominant singular value $\mathbf{s}_{1}$ contributes most to the score, removing $\mathbf{s}_{1}$ is likely to make many samples have similar scores. This would concentrate samples in a smaller region and further skew the distribution. Given that the OOD feature tends to have a much larger $\mathbf{s}_{1}$, this would have a greater impact on OOD data and skew the OOD score distribution more.

\begin{figure}[htbp]
    \centering
    \includegraphics[width=0.99\linewidth]{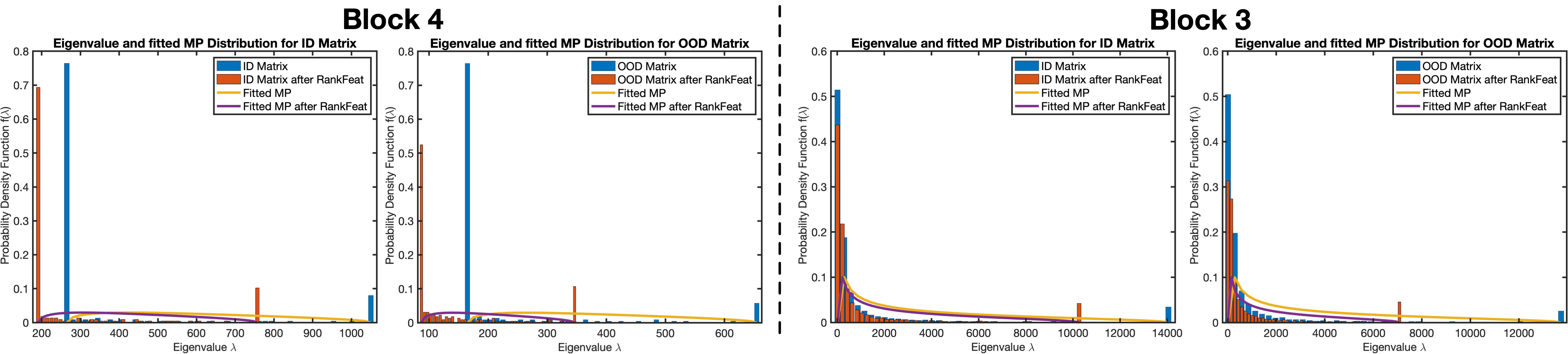}
    \caption{The exemplary eigenvalue distribution of ID/OOD feature and the fitted MP distribution. After the rank-1 matrix is removed, the lowest bin of OOD feature has a larger reduction and the middle bins gain some growth, making the ODD feature statistics closer to the MP distribution. }
    \label{fig:mp_dist}
\end{figure}

\noindent \textbf{Removing the rank-1 matrix is likely to make the statistics of OOD features closer to random matrices.} Now we turn to use RMT to analyze the statistics of OOD and ID feature matrices. For a random matrix of a given shape, the density of its eigenvalue asymptotically converges to the Manchenko-Pastur (MP) distribution~\cite{marvcenko1967distribution,sengupta1999distributions}. Formally, we have:
\begin{thm}[Manchenko-Pastur Law~\cite{marvcenko1967distribution,sengupta1999distributions}]
Let $\mathbf{X}$ be a random matrix of shape $t{\times}n$ whose entries are random variables with $E(\mathbf{X}_{ij}=0)$ and $E(\mathbf{X}_{ij}^2=1)$. Then the eigenvalues of the sample covariance $\mathbf{Y}=\frac{1}{n}\mathbf{X}\mathbf{X}^{T}$ converges to the probability density function: $\rho(\lambda) = \frac{t}{n} \frac{\sqrt{(\lambda_{+}-\lambda)(\lambda-\lambda_{-})}}{2\pi\lambda\sigma^2}\ for\  \lambda\in[\lambda_{-},\lambda_{+}]$ where $\lambda_{-}{=}\sigma^{2} (1-\sqrt{\frac{n}{t}})^2$ and $ \lambda_{+}{=}\sigma^{2} (1+\sqrt{\frac{n}{t}})^2$.
\end{thm}

This theorem implies the possibility to measure the statistical distance between ID/OOD features and random matrices. To this end, we randomly sample $1,000$ ID and OOD feature matrices and compute the KL divergence between the actual eigenvalue distribution and the fitted MP distribution.


\begin{table}[htbp]
    \caption{The KL divergence between ID/OOD feature and the fitted MP distribution. When the rank-1 feature is removed, the statistics of OOD matrix are closer to random matrices. }
    \centering
    \resizebox{0.7\linewidth}{!}{
    \begin{tabular}{c|cc|cc}
    \toprule
        \multirow{2}*{\textbf{Matrix Type}} & \multicolumn{2}{c|}{\textbf{Block 4}} & \multicolumn{2}{c}{\textbf{Block 3}} \\
        \cmidrule{2-5}
         & ID & OOD & ID & OOD \\
         \midrule
         Original feature matrix & 18.36 & 18.24 & 11.27 & 11.18\\
         Removing rank-1 matrix & 17.07 ($\downarrow$ 1.29)& \textbf{15.79 ($\downarrow$ 2.45)}& 9.84 ($\downarrow$ 1.45) & \textbf{8.71 ($\downarrow$ 2.47)} \\
    \bottomrule
    \end{tabular}
    }
    \label{tab:mp_distance}
\end{table}

Fig.~\ref{fig:mp_dist} and Table~\ref{tab:mp_distance} present the exemplary eigenvalue distribution and the average evaluation results of Block 4 and Block 3 features, respectively. For the original feature, the OOD and ID feature matrices exhibit similar behaviors: the distances to the fitted MP distribution are roughly the same ($diff.{\approx}0.1$). However, when the rank-1 matrix is removed, the OOD feature matrix has a much larger drop in the KL divergence. This indicates that removing the rank-1 matrix makes the statistics of OOD feature closer to random matrices, \emph{i.e.,} the OOD feature is very likely to become less informative than the ID feature. The result partly explains the working mechanism of \texttt{RankFeat}: \textit{by removing the feature matrix where OOD data might convey more information than ID data, the two types of distributions have a larger discrepancy and therefore can be better separated.}


\noindent \textbf{Connection with ReAct~\cite{sun2021react}.} \texttt{ReAct} clips the activations at the penultimate layer of a model to distinguish ID and OOD samples. Given the feature $\mathbf{X}$ and the pooling layer $\mathbf{m}$, the perturbation can be defined as:
\begin{equation}
     \min(\mathbf{X}\mathbf{m},\tau) = \mathbf{X}\mathbf{m} - \max(\mathbf{X}\mathbf{m}-\tau,0) 
\end{equation}
where $\tau$ is a pre-defined threshold. Their method shares some similarity with \texttt{RankFeat} formulation $\mathbf{X}\mathbf{m}{-}\mathbf{s}_{1}\mathbf{u}_{1}\mathbf{v}_{1}^{T}\mathbf{m}$. \textit{Both approaches subtract from the feature a portion of information that is most likely to cause the over-confidence of OOD prediction.} \texttt{ReAct} selects the manually-defined threshold $\tau$ based on statistics of the whole ID set, while \texttt{RankFeat} generates the structured rank-1 matrix from the feature itself. Taking a step further, \texttt{ReAct} has the score inequality following eq.~\eqref{eq:score_upper}
\begin{equation}
    \texttt{ReAct}(\mathbf{x})<  || \mathbf{W}\mathbf{X}\mathbf{m} - \mathbf{W}\max(\mathbf{X}\mathbf{m}-\tau,0) ||_{\infty} + ||\mathbf{b}||_{\infty} + \log(Q)
\end{equation}
Since $\mathbf{X}$ is non-negative (output of \texttt{ReLU}),
we have $\max(\mathbf{X}\mathbf{m})\geq\nicefrac{\max(\mathbf{X})}{HW}$. Exploiting the vector norm inequality $||\mathbf{X}||_{\rm F}{\geq}||\mathbf{X}||_{2}$ leads to the relation $\max(\mathbf{X}){\geq}\nicefrac{\mathbf{s}_{1}}{\sqrt{CHW}}$. Relying on this property, the above inequality can be re-formulated as:
\begin{equation}
    \texttt{ReAct}(\mathbf{x})< \frac{1}{HW}\sum_{i=1}^{N} \mathbf{s}_{i} ||\mathbf{W}||_{\infty} - \boxed{\frac{1}{HW}{\max( \frac{\mathbf{s}_{1} }{\sqrt{CHW}}-\tau, 0 )||\mathbf{W}||_{\infty}}}+ ||\mathbf{b}||_{\infty} + \log(Q)
\end{equation}
As indicated above, the upper bound of \texttt{ReAct} is also determined by the largest singular value $\mathbf{s}_{1}$. In contrast, the upper bound of our \texttt{RankFeat} can be expressed as:
\begin{equation}
    \texttt{RankFeat}(\mathbf{x})< \frac{1}{HW}\sum_{i=1}^{N} \mathbf{s}_{i} ||\mathbf{W}||_{\infty} - \boxed{\frac{1}{HW}\mathbf{s}_{1} ||\mathbf{W}||_{\infty}} + ||\mathbf{b}||_{\infty} + \log(Q)
\end{equation}
The upper bounds of both methods resemble each other with the only different term boxed. \textit{From this point of view, both methods distinguish the ID and OOD data by eliminating the impact of the term containing $\mathbf{s}_{1}$ in the upper bound.} \texttt{ReAct} optimizes it by clipping the term with a manually-defined threshold, which is indirect and might be sub-optimal. Moreover, the threshold selection requires statistics of the whole ID set. In contrast, our \texttt{RankFeat} does not require any extra data and directly subtracts this underlying term which is likely to cause the over-confidence of OOD samples.

\section{Experimental Results}

In this section, we first discuss the setup in Sec.~\ref{sec:exp_setup}, and then present the main experimental results on ImageNet-1k in Sec.~\ref{sec:exp_result}, 
followed by the extensive ablation studies in Sec.~\ref{sec:exp_ablation}.

\subsection{Setup}
\label{sec:exp_setup}

\noindent \textbf{Datasets.} In line with~\cite{huang2021mos,sun2021react,huang2021importance}, we mainly evaluate our method on the large-scale ImageNet-1k benchmark~\cite{deng2009imagenet}. The large-scale dataset is more challenging than the traditional CIFAR benchmark~\cite{krizhevsky2009learning} because the images are more realistic and diverse (\emph{i.e.,} $1.28$M images of $1,000$ classes). For the OOD datasets, we select four testsets from subsets of \texttt{iNaturalist}~\cite{van2018inaturalist}, \texttt{SUN}~\cite{xiao2010sun}, \texttt{Places}~\cite{zhou2017places}, and \texttt{Textures}~\cite{cimpoi2014describing}. These datasets are crafted by~\cite{huang2021mos} with non-overlapping categories from ImageNet-1k. Besides the experiment on the large-scale benchmark, we also validate the effectiveness of our approach on Species~\cite{hendrycks2019scaling} and CIFAR~\cite{krizhevsky2009learning} benchmark. (see Supplementary Material).


\noindent \textbf{Baselines.}
We compare our method with $6$ recent \emph{post hoc} OOD detection methods, namely \texttt{MSP}~\cite{hendrycks2016baseline}, \texttt{ODIN}~\cite{liang2017enhancing}, \texttt{Energy}~\cite{liu2020energy}, \texttt{Mahalanobis}~\cite{lee2018simple}, \texttt{GradNorm}~\cite{huang2021importance}, and \texttt{ReAct}~\cite{sun2021react}. The detailed illustration and settings of these methods are kindly referred to Supplementary Material.

\noindent \textbf{Architectures.} In line with~\cite{huang2021importance}, the main evaluation is done using Google BiT-S model~\cite{kolesnikov2020big} pretrained on ImageNet-1k with ResNetv2-101~\cite{he2016identity}. We also evaluate the performance on SqueezeNet~\cite{iandola2016squeezenet}, an alternative tiny architecture suitable for mobile devices and on T2T-ViT-24~\cite{yuan2021tokens}, a tokens-to-tokens vision transformer that has impressive performance when trained from scratch. 

\textit{For the implementation details and evaluation metrics, please refer to Supplementary Material}.

\begin{table}[htbp]
    \caption{Main evaluation results on ResNetv2-101~\cite{he2016identity}. All values are reported in percentages, and these \emph{post hoc} methods are directly applied to the model pre-trained on ImageNet-1k~\cite{deng2009imagenet}. The best three results are highlighted with \textbf{\textcolor{red}{red}}, \textbf{\textcolor{blue}{blue}}, and \textbf{\textcolor{cyan}{cyan}}.}
    \centering
    \resizebox{0.99\linewidth}{!}{
    \begin{tabular}{c|cc|cc|cc|cc|cc}
    \toprule
        \multirow{3}*{\textbf{Methods}} & \multicolumn{2}{c|}{\textbf{iNaturalist}} & \multicolumn{2}{c|}{\textbf{SUN}} & \multicolumn{2}{c|}{\textbf{Places}} & \multicolumn{2}{c|}{\textbf{Textures}} & \multicolumn{2}{c}{\textbf{Average}}   \\
        \cmidrule{2-11}
         & FPR95 & AUROC  & FPR95 & AUROC  & FPR95 & AUROC  & FPR95 & AUROC & FPR95 & AUROC  \\
         & ($\downarrow$) & ($\uparrow$) & ($\downarrow$) & ($\uparrow$) & ($\downarrow$) & ($\uparrow$) & ($\downarrow$) & ($\uparrow$) & ($\downarrow$) & ($\uparrow$)\\
    \midrule
    MSP~\cite{hendrycks2016baseline} & 63.69 & 87.59 & 79.89 & 78.34 & 81.44 & 76.76 & 82.73 & 74.45 & 76.96 & 79.29\\
    ODIN~\cite{liang2017enhancing} & 62.69 & 89.36 & 71.67 & 83.92 & 76.27 & 80.67 & 81.31 & 76.30 & 72.99 & 82.56\\
    Energy~\cite{liu2020energy} & 64.91 & 88.48 & 65.33 & 85.32 & 73.02 & 81.37 & 80.87 & 75.79 & 71.03 & 82.74\\
    Mahalanobis~\cite{lee2018simple} &96.34 &46.33 &88.43 & 65.20 &89.75 &64.46 &52.23 &72.10 &81.69 &62.02\\
    GradNorm~\cite{huang2021importance} &50.03 &90.33 &46.48 &89.03 &60.86 &84.82 &61.42 &81.07 &54.70 &86.71\\
    ReAct~\cite{sun2021react} &\textbf{\textcolor{blue}{44.52}}&\textbf{\textcolor{blue}{91.81}}&52.71&90.16&62.66&87.83&70.73&76.85&57.66&86.67\\
    \midrule
    \rowcolor{gray!20}\textbf{RankFeat (Block 4)} &\textbf{\textcolor{cyan}{46.54}} &81.49 &\textbf{\textcolor{red}{27.88}} & \textbf{\textcolor{blue}{92.18}} &\textbf{\textcolor{red}{38.26}} & \textbf{\textcolor{blue}{88.34}} & \textbf{\textcolor{cyan}{46.06}} & \textbf{\textcolor{cyan}{89.33}} & \textbf{\textcolor{blue}{39.69}} &\textbf{\textcolor{cyan}{87.84}}\\
    \rowcolor{gray!20}\textbf{RankFeat (Block 3)} &{49.61} &\textbf{\textcolor{cyan}{91.42}} &\textbf{\textcolor{cyan}{39.91}} &\textbf{\textcolor{cyan}{92.01}} &\textbf{\textcolor{cyan}{51.82}} &\textbf{\textcolor{cyan}{88.32}} &\textbf{\textcolor{blue}{41.84}} &\textbf{\textcolor{blue}{91.44}} &\textbf{\textcolor{cyan}{45.80}} &\textbf{\textcolor{blue}{90.80}} \\
     \rowcolor{gray!20} \textbf{RankFeat (Block 3 + 4)}
     &\textbf{\textcolor{red}{41.31}}&\textbf{\textcolor{red}{91.91}}&\textbf{\textcolor{blue}{29.27}}&\textbf{\textcolor{red}{94.07}}&\textbf{\textcolor{blue}{39.34}}&\textbf{\textcolor{red}{90.93}}&\textbf{\textcolor{red}{37.29}}&\textbf{\textcolor{red}{91.70}}&\textbf{\textcolor{red}{36.80}}&\textbf{\textcolor{red}{92.15}}\\
    \bottomrule
    \end{tabular}
    }
    \label{tab:main_results_res101}
\end{table}

\subsection{Results}
\label{sec:exp_result}

\noindent \textbf{Main results.} Following~\cite{huang2021importance}, the main evaluation is conducted using Google BiT-S model~\cite{kolesnikov2020big} pretrained on ImageNet-1k with ResNetv2-101 architecture~\cite{he2016identity}. Table~\ref{tab:main_results_res101} compares the performance of all the \emph{post hoc} methods. For both Block 3 and Block 4 features, our \texttt{RankFeat} achieves the best evaluation results across datasets and metrics. More specifically, \texttt{RankFeat} based on the Block 4 feature outperforms the second-best baseline by $\textbf{15.01\%}$ in the average FPR95, while the Block 3 feature-based \texttt{RankFeat} beats the second-best method by $\textbf{4.09\%}$ in the average AUROC. Their combination further surpasses other methods by $\textbf{17.90\%}$ in the average FPR95 and by $\textbf{5.44\%}$ in the average AUROC. The superior performances at various depths demonstrate the effectiveness and general applicability of \texttt{RankFeat}. The Block 3 feature has a higher AUROC but slightly falls behind the Block 4 feature in the FPR95, which can be considered a compromise between the two metrics.


\begin{table}[htbp]
    \caption{The results on SqueezeNet~\cite{iandola2016squeezenet} and T2T-ViT-24~\cite{yuan2021tokens}. All values are reported in percentages, and these \emph{post hoc} methods are directly applied to the model pre-trained on ImageNet-1k~\cite{deng2009imagenet}. For results on SqueezeNet~\cite{iandola2016squeezenet}, the best three results are highlighted with \textbf{\textcolor{red}{red}}, \textbf{\textcolor{blue}{blue}}, and \textbf{\textcolor{cyan}{cyan}}.}
    \centering
    \resizebox{0.99\linewidth}{!}{
    \begin{tabular}{c|c|cc|cc|cc|cc|cc}
    \toprule
        \multirow{3}*{\textbf{Model}}&\multirow{3}*{\textbf{Methods}} & \multicolumn{2}{c|}{\textbf{iNaturalist}} & \multicolumn{2}{c|}{\textbf{SUN}} & \multicolumn{2}{c|}{\textbf{Places}} & \multicolumn{2}{c|}{\textbf{Textures}} & \multicolumn{2}{c}{\textbf{Average}}   \\
        \cmidrule{3-12}
         && FPR95 & AUROC  & FPR95 & AUROC  & FPR95 & AUROC  & FPR95 & AUROC & FPR95 & AUROC  \\
         && ($\downarrow$) & ($\uparrow$) & ($\downarrow$) & ($\uparrow$) & ($\downarrow$) & ($\uparrow$) & ($\downarrow$) & ($\uparrow$) & ($\downarrow$) & ($\uparrow$)\\
    \midrule
    \multirow{9}*{\textbf{SqueezeNet~\cite{iandola2016squeezenet}}}&MSP~\cite{hendrycks2016baseline} 
    &89.83&65.41&83.03&72.25&87.27&67.00&94.61&41.84&88.84&61.63\\
    &ODIN~\cite{liang2017enhancing} &90.79&65.75&78.32&78.37&83.23&73.31&92.25&43.43&86.15&65.17\\
    &Energy~\cite{liu2020energy} &79.27&73.30&56.41&87.88&67.74&82.73&67.16&64.51&67.65&77.11\\
    &Mahalanobis~\cite{lee2018simple} &91.50&51.79&90.33&62.18&92.26&56.63&58.60&67.16&83.17&59.44\\
    &GradNorm~\cite{huang2021importance} &76.31&73.92&53.63&87.55&65.99&\textbf{\textcolor{cyan}{83.28}}&68.72&68.07&66.16&78.21\\
    &ReAct~\cite{sun2021react} &76.78&68.56&87.57&66.37&88.80&66.20&51.05&76.57&76.05&69.43\\
    \cmidrule{2-12}
   \rowcolor{gray!20} \cellcolor{white}&\textbf{RankFeat (Block 4)} &\textbf{\textcolor{red}{61.67}}&\textbf{\textcolor{red}{83.09}}&\textbf{\textcolor{blue}{46.72}}&\textbf{\textcolor{cyan}{88.31}}&\textbf{\textcolor{red}{61.31}}&{80.52}&\textbf{\textcolor{red}{38.04}}&\textbf{\textcolor{red}{88.82}}&\textbf{\textcolor{red}{51.94}}&\textbf{\textcolor{blue}{85.19}}\\
    \rowcolor{gray!20} \cellcolor{white}&\textbf{RankFeat (Block 3)} &\textbf{\textcolor{cyan}{71.04}}&\textbf{\textcolor{cyan}{81.50}}&\textbf{\textcolor{cyan}{49.18}}&\textbf{\textcolor{red}{90.43}}&\textbf{\textcolor{cyan}{62.94}}&\textbf{\textcolor{red}{85.82}}&\textbf{\textcolor{cyan}{50.14}}&\textbf{\textcolor{cyan}{79.32}}&\textbf{\textcolor{cyan}{58.33}}&\textbf{\textcolor{cyan}{84.28}} \\
    \rowcolor{gray!20} \cellcolor{white}&\textbf{RankFeat (Block 3 + 4)}
    &\textbf{\textcolor{blue}{65.81}}&\textbf{\textcolor{blue}{83.06}}&\textbf{\textcolor{red}{46.64}}&\textbf{\textcolor{blue}{90.17}}&\textbf{\textcolor{blue}{61.56}}&\textbf{\textcolor{blue}{84.51}}&\textbf{\textcolor{blue}{42.54}}&\textbf{\textcolor{blue}{85.00}}&\textbf{\textcolor{blue}{54.14}}&\textbf{\textcolor{red}{85.69}}\\
    \midrule
    \multirow{7}*{\textbf{T2T-ViT-24~\cite{yuan2021tokens}}}&MSP~\cite{hendrycks2016baseline} &48.92&88.95&61.77&81.37&69.54&80.03&62.91&82.31&60.79&83.17\\ 
    &ODIN~\cite{liang2017enhancing} &\textbf{44.07}&88.17&63.83&78.46&68.19&75.33&54.27&83.63&57.59&81.40\\
    &Energy~\cite{liu2020energy} &52.95&82.93&68.55&73.06&74.24&68.17&51.05&83.25&61.70&76.85\\
    &Mahalanobis~\cite{lee2018simple} &90.50&58.13&91.71&50.52&93.32&49.60&80.67&64.06&89.05&55.58\\
    &GradNorm~\cite{huang2021importance} &99.30&25.86&98.37&28.06&99.01&25.71&92.68&38.80&97.34&29.61\\
    &ReAct~\cite{sun2021react} &52.17&\textbf{89.51}&65.23&81.03&68.93&78.20&52.54&85.46&59.72&83.55\\
    \cmidrule{2-12}
   \rowcolor{gray!20} \cellcolor{white}&\textbf{RankFeat} &50.27 & 87.81 & \textbf{57.18} & \textbf{84.33} & \textbf{66.22} & \textbf{80.89} & \textbf{32.64} & \textbf{89.36} &\textbf{51.58} & \textbf{85.60} \\
    \bottomrule
    \end{tabular}
    }
    \label{tab:main_results_vit}
\end{table}

\noindent \textbf{Our RankFeat is also effective on alternative CNN architectures.}
Besides the experiment on ResNetv2~\cite{he2016identity}, we also evaluate our method on SqueezeNet~\cite{iandola2016squeezenet}, an alternative tiny network suitable for mobile devices and on-chip applications. This network is more challenging because the tiny network size makes the model prone to overfit the training data, which could increase the difficulty to distinguish between ID and OOD samples. Table~\ref{tab:main_results_vit} top presents the performance of all the methods. Collectively, the performances of \texttt{RankFeat} are very competitive at both depths, as well as the score fusion. Our \texttt{RankFeat} achieves the \emph{state-of-the-art} performances, outperforming the second-best baseline by $\textbf{14.22\%}$ in FPR95 and by $\textbf{7.48\%}$ in AUROC.

\noindent \textbf{Our RankFeat also suits transformer-based architectures.} To further demonstrate the applicability of our method, we evaluate \texttt{RankFeat} on Tokens-to-Tokens Vision Transformer (T2T-ViT)~\cite{yuan2021tokens}, a popular transformer architecture that can achieve competitive performance against CNNs when trained from scratch. Similar to the CNN, \texttt{RankFeat} removes the rank-1 matrix from the final token of T2T-ViT before the last normalization layer and the classification head. Table~\ref{tab:main_results_vit} bottom compares the performance on T2T-ViT-24. Our \texttt{RankFeat} outperforms the second-best method by $\textbf{6.11\%}$ in FPR95 and by $\textbf{2.05\%}$ in AUROC. Since the transformer models~\cite{dosovitskiy2020image,yuan2021tokens} do not have increasing receptive fields like CNNs, we do not evaluate the performance at alternative network depths.

\noindent \textbf{Comparison against training-needed approaches.} Since our method is \emph{post hoc}, we only compare it with other \emph{post hoc} baselines. MOS~\cite{huang2021mos} and KL Matching~\cite{hendrycks2019scaling} are not taken into account because MOS needs extra training processes and KL Matching requires the labeled validation set to compute distributions for each class. Nonetheless, we note that our method can still hold an advantage against those approaches. Table~\ref{tab:mos_kl} presents the average FPR95 and AUROC on the ImageNet-1k benchmark. Our RankFeat achieves the best performance without any extra training or validation set.

\begin{table}[htbp]
    \centering
    \caption{Comparison against training-needed methods on ImageNet-1k based on ResNetv2-101~\cite{he2016identity}.}
    \resizebox{0.7\linewidth}{!}{
    \begin{tabular}{c|c|c|c|c}
    \toprule
         Method & Post hoc? & Free of Validation Set? & FPR95 ($\downarrow$) & AUROC ($\uparrow$)\\
    \midrule
         KL Matching~\cite{hendrycks2019scaling}& $\textcolor{green}{\cmark}$ &$\textcolor{red}{\xmark}$&54.30&80.82 \\
         MOS~\cite{huang2021mos} & $\textcolor{red}{\xmark}$ & $\textcolor{green}{\cmark}$ &39.97&90.11 \\
         \rowcolor{gray!20}\textbf{RankFeat} & $\textcolor{green}{\cmark}$ & $\textcolor{green}{\cmark}$ & \textbf{36.80} &\textbf{92.15}\\
    \bottomrule
    \end{tabular}
    }
    \label{tab:mos_kl}
\end{table}

\subsection{Ablation Studies}
\label{sec:exp_ablation}

In this subsection, we conduct several ablation studies based on Google-BiT-S ResNetv2-101 model. Unless explicitly specified, we apply \texttt{RankFeat} on the Block 4 feature by default. 

\begin{table}[htbp]
    \caption{Ablation studies on keeping only the rank-1 matrix and removing the rank-n matrix. }
    \centering
    \resizebox{0.99\linewidth}{!}{
    \begin{tabular}{c|cc|cc|cc|cc|cc}
    \toprule
        \multirow{3}*{\textbf{Baselines}} & \multicolumn{2}{c|}{\textbf{iNaturalist}} & \multicolumn{2}{c|}{\textbf{SUN}} & \multicolumn{2}{c|}{\textbf{Places}} & \multicolumn{2}{c|}{\textbf{Textures}} & \multicolumn{2}{c}{\textbf{Average}}   \\
        \cmidrule{2-11}
         & FPR95 & AUROC  & FPR95 & AUROC  & FPR95 & AUROC  & FPR95 & AUROC & FPR95 & AUROC  \\
         & ($\downarrow$) & ($\uparrow$) & ($\downarrow$) & ($\uparrow$) & ($\downarrow$) & ($\uparrow$) & ($\downarrow$) & ($\uparrow$) & ($\downarrow$) & ($\uparrow$)\\
    \midrule
    GradNorm~\cite{huang2021importance} &50.03 &90.33 &46.48 &89.03 &60.86 &84.82 &61.42 &81.07 &54.70 &86.71\\
    ReAct~\cite{sun2021react} &\textbf{44.52}&{91.81}&52.71&90.16&62.66&87.83&70.73&76.85&57.66&86.67\\
    \midrule
    Keeping Only Rank-1 &48.97&\textbf{91.93}&62.63&84.62&72.42&79.79&49.42&88.86&58.49&86.30 \\
    Removing Rank-3 &55.19&90.03&48.97&91.26&56.63&\textbf{88.81}&86.95&74.57&61.94&86.17\\
    Removing Rank-2 &50.04&89.30&48.55&90.99&56.23&88.38&76.86&81.37&57.92&87.51\\
    \midrule
    \rowcolor{gray!20}\textbf{Removing Rank-1} &\textbf{46.54} &81.49 &\textbf{27.88} & \textbf{92.18} &\textbf{38.26} & {88.34} & \textbf{46.06} & \textbf{89.33} & \textbf{39.69} &\textbf{87.84}\\
    \bottomrule
    \end{tabular}
    }
    \label{tab:ablation_rank}
\end{table}

\noindent \textbf{Removing the rank-1 matrix outperforms keeping only it.} Instead of removing the rank-1 matrix, another seemingly promising approach is keeping only the rank-1 matrix and abandoning the rest of the matrix. Table~\ref{tab:ablation_rank} presents the evaluation results of keeping only the rank-1 matrix. The performance falls behind that of removing the rank-1 feature by $18.8\%$ in FPR95, which indicates that keeping only the rank-1 feature is inferior to removing it in distinguishing the two distributions. Nonetheless, it is worth noting that even keeping only the rank-1 matrix achieves very competitive performance against previous best methods, such as \texttt{GradNorm}~\cite{huang2021importance} and \texttt{ReAct}~\cite{sun2021react}.

\noindent \textbf{Removing the rank-1 matrix outperforms removing the rank-n matrix (n>1).} We evaluate the impact of removing the matrix of a higher rank, \emph{i.e.,} performing $\mathbf{X}{-}\sum_{i=1}^{n}\mathbf{s}_{i}\mathbf{u}_{i}\mathbf{v}_{i}^{T}$ where $n{>}1$ for the high-level feature $\mathbf{X}$. Table~\ref{tab:ablation_rank} compares the performance of removing the rank-2 matrix and rank-3 matrix. When the rank of the removed matrix is increased, the average performance degrades accordingly. This demonstrates that removing the rank-1 matrix is the most effective approach to separate ID and OOD data. This result is coherent with the finding in Fig.~\ref{fig:cover}(a): only the largest singular value of OOD data is significantly different from that of ID data. Therefore, removing the rank-1 matrix achieves the best performance.  

\begin{table}[htbp]
    \caption{The ablation study on applying \texttt{RankFeat} to features at different network depths.} 
    \centering
    \resizebox{0.9\linewidth}{!}{
    \begin{tabular}{c|cc|cc|cc|cc|cc}
    \toprule
        \multirow{3}*{\textbf{Layer}} & \multicolumn{2}{c|}{\textbf{iNaturalist}} & \multicolumn{2}{c|}{\textbf{SUN}} & \multicolumn{2}{c|}{\textbf{Places}} & \multicolumn{2}{c|}{\textbf{Textures}} & \multicolumn{2}{c}{\textbf{Average}}   \\
        \cmidrule{2-11}
         & FPR95 & AUROC  & FPR95 & AUROC  & FPR95 & AUROC  & FPR95 & AUROC & FPR95 & AUROC  \\
         & ($\downarrow$) & ($\uparrow$) & ($\downarrow$) & ($\uparrow$) & ($\downarrow$) & ($\uparrow$) & ($\downarrow$) & ($\uparrow$) & ($\downarrow$) & ($\uparrow$)\\
    \midrule
    Block 1 &87.81&77.00&59.15&87.29&65.50&84.35&94.15&60.41&76.65&77.26\\
    Block 2 &71.84&85.80&61.44&86.46&71.68&81.65&87.89&72.04&73.23&81.49\\
    \rowcolor{gray!20}\textbf{Block 3} &\textbf{49.61} &\textbf{91.42} &\textbf{39.91} &\textbf{92.01} &\textbf{51.82} &\textbf{88.32} &\textbf{41.84} &\textbf{91.44} &\textbf{45.80} &\textbf{90.80} \\
    \rowcolor{gray!20}\textbf{Block 4} &\textbf{46.54} &81.49 &\textbf{27.88} & \textbf{92.18} &\textbf{38.26} & \textbf{88.34} & \textbf{46.06} & \textbf{89.33} & \textbf{39.69} &\textbf{87.84}\\
    \bottomrule
    \end{tabular}
    }
    \label{tab:ablation_block}
\end{table}

\noindent \textbf{Block 3 and Block 4 features are the most informative.} In addition to exploring the high-level features at Block 3 and Block 4, we also investigate the possibility of applying \texttt{RankFeat} to features at shallow network layers. As shown in Table~\ref{tab:ablation_block}, the performances of \texttt{RankFeat} at the Block 1 and Block 2 features are not comparable to those at deeper layers. This is mainly because the shallow low-level features do not embed as rich semantic information as the deep features. Consequently, removing the rank-1 matrix of shallow features would not help to separate the ID and OOD data. 




\begin{table}[htbp]
    \caption{The approximate solution by PI yields competitive performance and costs much less time consumption. The test batch size is set as $16$.}
    \centering
    \resizebox{0.99\linewidth}{!}{
    \begin{tabular}{c|c|cc|cc|cc|cc|cc}
    \toprule
        \multirow{3}*{\makecell[c]{\textbf{Computation} \\ \textbf{Technique}}} &
        \multirow{3}*{\makecell[c]{\textbf{Processing Time} \\ \textbf{Per Image} \\ \textbf{(ms)}}} 
        & \multicolumn{2}{c|}{\textbf{iNaturalist}} & \multicolumn{2}{c|}{\textbf{SUN}} & \multicolumn{2}{c|}{\textbf{Places}} & \multicolumn{2}{c|}{\textbf{Textures}} & \multicolumn{2}{c}{\textbf{Average}}   \\
        \cmidrule{3-12}
         & & FPR95 & AUROC  & FPR95 & AUROC  & FPR95 & AUROC  & FPR95 & AUROC & FPR95 & AUROC  \\
         & & ($\downarrow$) & ($\uparrow$) & ($\downarrow$) & ($\uparrow$) & ($\downarrow$) & ($\uparrow$) & ($\downarrow$) & ($\uparrow$) & ($\downarrow$) & ($\uparrow$)\\
         \midrule
          GradNorm~\cite{huang2021importance} &80.01
          &50.03 &90.33 &46.48 &89.03 &60.86 &84.82 &61.42 &81.07 &54.70 &86.71\\
          ReAct~\cite{sun2021react} &8.79 &\textbf{44.52}&\textbf{91.81}&52.71&90.16&62.66&87.83&70.73&76.85&57.66&86.67\\
          \midrule
         SVD & 18.01 &{46.54} &81.49 &\textbf{27.88} & \textbf{92.18} &{38.26} & \textbf{88.34} & \textbf{46.06} & \textbf{89.33} & \textbf{39.69} &\textbf{87.84}\\
         PI ($\#100$ iter) &9.97 &46.59&81.49&27.93&92.18&38.28&88.34&46.09&89.33&39.72&\textbf{87.84}\\
         PI ($\#50$ iter) &9.47 &46.58&81.49&27.93&92.17&\textbf{38.24}&88.34&46.12&89.32 &39.72&87.83\\
         PI ($\#20$ iter) &9.22 &46.58 &81.48 &27.93 &92.15 &38.28 &88.31 &46.10 &89.33 &39.75 &87.82 \\
         PI ($\#10$ iter) &9.03 &46.77&81.29&28.21&91.84&38.44&87.94&46.08&89.37&39.88&87.61 \\ 
         PI ($\#5$ iter) &9.00&48.34&79.81&30.44&89.71&41.33&84.97&45.34&89.41&41.36&85.98 \\ 
    \bottomrule
    \end{tabular}
    }
    \label{tab:ablation_approximate}
\end{table}

\noindent \textbf{The approximate solution by PI yields competitive performances.} Table~\ref{tab:ablation_approximate} compares the time consumption and performance of SVD and PI, as well as two recent \emph{state-of-the-art} OOD methods \texttt{ReAct} and \texttt{GradNorm}. The performance of PI starts to become competitive against that of SVD (${<}0.1\%$) from $\textbf{20}$ iterations on with $\textbf{48.41\%}$ time reduction. Compared with \texttt{ReAct}, the PI-based \texttt{RankFeat} only requires marginally $4.89\%$ more time consumption. \texttt{GradNorm} is not comparable against other baselines in terms of time cost because it does not support the batch mode.

\begin{figure}[htbp]
    \centering
    \includegraphics[width=0.8\linewidth]{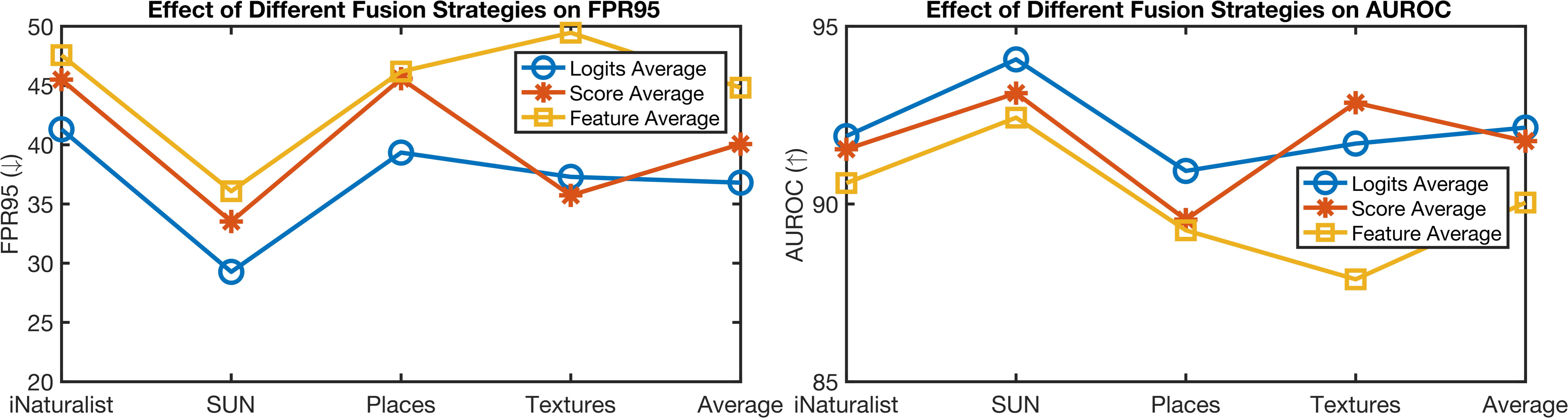}
    \caption{The impact of different fusion strategies on FPR95 and AUROC.}
    \label{fig:fusion}
\end{figure}

\noindent \textbf{Fusion at the logit space achieves the best performance.} Fig.~\ref{fig:fusion} displays the performance of different fusion strategies in combining \texttt{RankFeat} at the Block 3 and Block 4 features. As can be observed, averaging logits outperforms other fusion strategies in most datasets and metrics. This indicates that the fusing the logits can best coordinate the benefit of both features.   

\section{Related Work}

\noindent \textbf{Distribution shifts.} Distribution shifts have been a long-standing problem in the machine learning research community~\cite{hand2006classifier,quinonero2008dataset,koh2021wilds,wiles2021fine}. The problem of distributions shifts can be generally categorized as shifts in the input space and shifts in the label space. Shifts only in the input space are often deemed as \emph{covariate shifts}~\cite{hendrycks2018benchmarking,ovadia2019can}. In this setting, the inputs are corrupted by perturbations or shifted by domains, but the label space stays the same~\cite{hsu2020generalized,sun2020test}. The aim is mainly to improve the robustness and generalization of a model~\cite{hendrycks2019augmix}. For OOD detection, the labels are disjoint and the main concern is to determine whether a test sample should be predicted by the pre-trained model~\cite{liang2017enhancing,hsu2020generalized}. 

Some related sub-fields also tackle the problem of distribution shifts in the label space, such as novel class discovery~\cite{han2019learning,zhong2021neighborhood}, open-set recognition~\cite{scheirer2012toward,vaze2021open}, and novelty detection~\cite{abati2019latent,tack2020csi}. These sub-fields target specific distribution shifts (\emph{e.g.,} semantic novelty), while OOD encompasses all forms of shifts.


\noindent \textbf{OOD detection with discriminative models.} 
The early work on discriminative OOD detection dates back to the classification model with rejection option~\cite{chow1970optimum,fumera2002support}. The OOD detection methods can be generally divided into training-need methods and \emph{post hoc} approaches. Compared with training-needed approaches, \emph{post hoc} methods do not require any extra training processes and could be directly applied to any pre-trained models. For the wide body of research on OOD detection, please refer to~\cite{yang2021generalized} for the comprehensive survey. Here we only highlight the representative \emph{post hoc} methods. Nguyen~\emph{et al.}~\cite{nguyen2015deep} first observed the phenomenon that neural networks easily give over-confident predictions for OOD samples. The following researches attempted to improve the OOD uncertainty estimation by proposing ODIN score~\cite{liang2017enhancing}, OpenMax score~\cite{bendale2016towards}, Mahalanobis distance~\cite{lee2018simple}, and Energy score~\cite{liu2020energy}.
Huang~\emph{et al.}~\cite{huang2021mos} pointed out that the traditional CIFAR benchmark does not extrapolate to real-world settings and proposed a large-scale ImageNet benchmark. More recently, Sun~\emph{et al.}~\cite{sun2021react} and Huang~\emph{et al.}~\cite{huang2021importance} proposed to tackle the challenge of OOD detection from the lens of activation abnormality and gradient norm, respectively. In contrast, based on the empirical observation of singular value distributions, we propose a simple yet effective \emph{post hoc} solution by removing the rank-1 subspace from the high-level features. 

\noindent \textbf{OOD detection with generative models.} Different from discriminative models, generative models detect the OOD samples by estimating the probability density function~\cite{kingma2013auto,tabak2013family,rezende2014stochastic,van2016conditional,dinh2016density,huang2017stacked,bibas2021single,jiang2021revisiting}. A sample with a low likelihood is deemed as OOD data. Recently, a multitude of methods have utilized generative models for OOD detection~\cite{ren2019likelihood,serra2019input,wang2020further,xiao2020likelihood,kirichenko2020normalizing,schirrmeister2020understanding,kim2021locally}. However, as pointed out in~\cite{nalisnick2018deep}, generative models could assign a high likelihood to OOD data. Furthermore, generative models can be prohibitively harder to train and optimize than their discriminative counterparts, and the performance is often inferior. This might limit their practical usage.  

\section{Conclusion}

In this paper, we present \texttt{RankFeat}, a simple yet effective approach for \emph{post hoc} OOD detection by removing the rank-1 matrix composed by the largest singular value from the high-level feature. We demonstrate its superior empirical results and the general applicability across architectures, network depths, and benchmarks. Extensive ablation studies and comprehensive theoretical analyses are conducted to reveal the important insights and to explain the working mechanism of our method.

\begin{ack}
This research was supported by the EU H2020 projects AI4Media (No. 951911) and SPRING (No. 871245). We thank our colleague Zhun Zhong for the fruitful discussion and valuable suggestions.
\end{ack}

{
\small
\bibliographystyle{plainnat}
\bibliography{egbib}

\begin{thebibliography}{74}
\providecommand{\natexlab}[1]{#1}
\providecommand{\url}[1]{\texttt{#1}}
\expandafter\ifx\csname urlstyle\endcsname\relax
  \providecommand{\doi}[1]{doi: #1}\else
  \providecommand{\doi}{doi: \begingroup \urlstyle{rm}\Url}\fi

\bibitem[Abati et~al.(2019)Abati, Porrello, Calderara, and
  Cucchiara]{abati2019latent}
Davide Abati, Angelo Porrello, Simone Calderara, and Rita Cucchiara.
\newblock Latent space autoregression for novelty detection.
\newblock In \emph{CVPR}, pages 481--490, 2019.

\bibitem[Bendale and Boult(2016)]{bendale2016towards}
Abhijit Bendale and Terrance~E Boult.
\newblock Towards open set deep networks.
\newblock In \emph{CVPR}, pages 1563--1572, 2016.

\bibitem[Bibas et~al.(2021)Bibas, Feder, and Hassner]{bibas2021single}
Koby Bibas, Meir Feder, and Tal Hassner.
\newblock Single layer predictive normalized maximum likelihood for
  out-of-distribution detection.
\newblock \emph{NeurIPS}, 34, 2021.

\bibitem[Chow(1970)]{chow1970optimum}
C~Chow.
\newblock On optimum recognition error and reject tradeoff.
\newblock \emph{IEEE Transactions on information theory}, 16\penalty0
  (1):\penalty0 41--46, 1970.

\bibitem[Cimpoi et~al.(2014)Cimpoi, Maji, Kokkinos, Mohamed, and
  Vedaldi]{cimpoi2014describing}
Mircea Cimpoi, Subhransu Maji, Iasonas Kokkinos, Sammy Mohamed, and Andrea
  Vedaldi.
\newblock Describing textures in the wild.
\newblock In \emph{CVPR}, 2014.

\bibitem[Deng et~al.(2009)Deng, Dong, Socher, Li, Li, and
  Fei-Fei]{deng2009imagenet}
Jia Deng, Wei Dong, Richard Socher, Li-Jia Li, Kai Li, and Li~Fei-Fei.
\newblock Imagenet: A large-scale hierarchical image database.
\newblock In \emph{CVPR}, pages 248--255. Ieee, 2009.

\bibitem[Diffenderfer et~al.(2021)Diffenderfer, Bartoldson, Chaganti, Zhang,
  and Kailkhura]{diffenderfer2021winning}
James Diffenderfer, Brian Bartoldson, Shreya Chaganti, Jize Zhang, and Bhavya
  Kailkhura.
\newblock A winning hand: Compressing deep networks can improve
  out-of-distribution robustness.
\newblock \emph{NeurIPS}, 34, 2021.

\bibitem[Ding et~al.(2021)Ding, Zhang, Ma, Han, Ding, and Sun]{ding2021repvgg}
Xiaohan Ding, Xiangyu Zhang, Ningning Ma, Jungong Han, Guiguang Ding, and Jian
  Sun.
\newblock Repvgg: Making vgg-style convnets great again.
\newblock In \emph{CVPR}, pages 13733--13742, 2021.

\bibitem[Dinh et~al.(2017)Dinh, Sohl-Dickstein, and Bengio]{dinh2016density}
Laurent Dinh, Jascha Sohl-Dickstein, and Samy Bengio.
\newblock Density estimation using real nvp.
\newblock \emph{ICLR}, 2017.

\bibitem[Dosovitskiy et~al.(2021)Dosovitskiy, Beyer, Kolesnikov, Weissenborn,
  Zhai, Unterthiner, Dehghani, Minderer, Heigold, Gelly,
  et~al.]{dosovitskiy2020image}
Alexey Dosovitskiy, Lucas Beyer, Alexander Kolesnikov, Dirk Weissenborn,
  Xiaohua Zhai, Thomas Unterthiner, Mostafa Dehghani, Matthias Minderer, Georg
  Heigold, Sylvain Gelly, et~al.
\newblock An image is worth 16x16 words: Transformers for image recognition at
  scale.
\newblock In \emph{ICLR}, 2021.

\bibitem[Du et~al.(2022)Du, Wang, Cai, and Li]{du2022vos}
Xuefeng Du, Zhaoning Wang, Mu~Cai, and Yixuan Li.
\newblock Vos: Learning what you don't know by virtual outlier synthesis.
\newblock \emph{ICLR}, 2022.

\bibitem[Emmott et~al.(2013)Emmott, Das, Dietterich, Fern, and
  Wong]{emmott2013systematic}
Andrew~F Emmott, Shubhomoy Das, Thomas Dietterich, Alan Fern, and Weng-Keen
  Wong.
\newblock Systematic construction of anomaly detection benchmarks from real
  data.
\newblock In \emph{Proceedings of the ACM SIGKDD workshop on outlier detection
  and description}, 2013.

\bibitem[Fort et~al.(2021)Fort, Ren, and Lakshminarayanan]{fort2021exploring}
Stanislav Fort, Jie Ren, and Balaji Lakshminarayanan.
\newblock Exploring the limits of out-of-distribution detection.
\newblock \emph{NeurIPS}, 34, 2021.

\bibitem[Fumera and Roli(2002)]{fumera2002support}
Giorgio Fumera and Fabio Roli.
\newblock Support vector machines with embedded reject option.
\newblock In \emph{International Workshop on Support Vector Machines}, pages
  68--82. Springer, 2002.

\bibitem[Garg et~al.(2022)Garg, Balakrishnan, Lipton, Neyshabur, and
  Sedghi]{garg2022leveraging}
Saurabh Garg, Sivaraman Balakrishnan, Zachary~C Lipton, Behnam Neyshabur, and
  Hanie Sedghi.
\newblock Leveraging unlabeled data to predict out-of-distribution performance.
\newblock \emph{ICLR}, 2022.

\bibitem[Gibbs and Candes(2021)]{gibbs2021adaptive}
Isaac Gibbs and Emmanuel Candes.
\newblock Adaptive conformal inference under distribution shift.
\newblock \emph{NeurIPS}, 34, 2021.

\bibitem[Golan and El-Yaniv(2018)]{golan2018deep}
Izhak Golan and Ran El-Yaniv.
\newblock Deep anomaly detection using geometric transformations.
\newblock \emph{NeurIPS}, 2018.

\bibitem[Gomes et~al.(2022)Gomes, Alberge, Duhamel, and
  Piantanida]{gomes2022igeood}
Eduardo Dadalto~Camara Gomes, Florence Alberge, Pierre Duhamel, and Pablo
  Piantanida.
\newblock Igeood: An information geometry approach to out-of-distribution
  detection.
\newblock \emph{ICLR}, 2022.

\bibitem[Goodfellow et~al.(2015)Goodfellow, Shlens, and
  Szegedy]{goodfellow2014explaining}
Ian~J Goodfellow, Jonathon Shlens, and Christian Szegedy.
\newblock Explaining and harnessing adversarial examples.
\newblock \emph{ICLR}, 2015.

\bibitem[Han et~al.(2019)Han, Vedaldi, and Zisserman]{han2019learning}
Kai Han, Andrea Vedaldi, and Andrew Zisserman.
\newblock Learning to discover novel visual categories via deep transfer
  clustering.
\newblock In \emph{ICCV}, pages 8401--8409, 2019.

\bibitem[Hand(2006)]{hand2006classifier}
David~J Hand.
\newblock Classifier technology and the illusion of progress.
\newblock \emph{Statistical science}, 21\penalty0 (1):\penalty0 1--14, 2006.

\bibitem[Haroush et~al.(2022)Haroush, Frostig, Heller, and
  Soudry]{haroush2021statistical}
Matan Haroush, Tzviel Frostig, Ruth Heller, and Daniel Soudry.
\newblock A statistical framework for efficient out of distribution detection
  in deep neural networks.
\newblock In \emph{ICLR}, 2022.

\bibitem[He et~al.(2016{\natexlab{a}})He, Zhang, Ren, and Sun]{he2016deep}
Kaiming He, Xiangyu Zhang, Shaoqing Ren, and Jian Sun.
\newblock Deep residual learning for image recognition.
\newblock In \emph{CVPR}, pages 770--778, 2016{\natexlab{a}}.

\bibitem[He et~al.(2016{\natexlab{b}})He, Zhang, Ren, and Sun]{he2016identity}
Kaiming He, Xiangyu Zhang, Shaoqing Ren, and Jian Sun.
\newblock Identity mappings in deep residual networks.
\newblock In \emph{ECCV}, pages 630--645. Springer, 2016{\natexlab{b}}.

\bibitem[Hendrycks and Dietterich(2019)]{hendrycks2018benchmarking}
Dan Hendrycks and Thomas Dietterich.
\newblock Benchmarking neural network robustness to common corruptions and
  perturbations.
\newblock In \emph{ICLR}, 2019.

\bibitem[Hendrycks and Gimpel(2017)]{hendrycks2016baseline}
Dan Hendrycks and Kevin Gimpel.
\newblock A baseline for detecting misclassified and out-of-distribution
  examples in neural networks.
\newblock \emph{ICLR}, 2017.

\bibitem[Hendrycks et~al.(2019)Hendrycks, Mu, Cubuk, Zoph, Gilmer, and
  Lakshminarayanan]{hendrycks2019augmix}
Dan Hendrycks, Norman Mu, Ekin~Dogus Cubuk, Barret Zoph, Justin Gilmer, and
  Balaji Lakshminarayanan.
\newblock Augmix: A simple data processing method to improve robustness and
  uncertainty.
\newblock In \emph{ICLR}, 2019.

\bibitem[Hendrycks et~al.(2022)Hendrycks, Basart, Mazeika, Mostajabi,
  Steinhardt, and Song]{hendrycks2019scaling}
Dan Hendrycks, Steven Basart, Mantas Mazeika, Mohammadreza Mostajabi, Jacob
  Steinhardt, and Dawn Song.
\newblock Scaling out-of-distribution detection for real-world settings.
\newblock \emph{ICML}, 2022.

\bibitem[Hsu et~al.(2020)Hsu, Shen, Jin, and Kira]{hsu2020generalized}
Yen-Chang Hsu, Yilin Shen, Hongxia Jin, and Zsolt Kira.
\newblock Generalized odin: Detecting out-of-distribution image without
  learning from out-of-distribution data.
\newblock In \emph{CVPR}, pages 10951--10960, 2020.

\bibitem[Huang and Li(2021)]{huang2021mos}
Rui Huang and Yixuan Li.
\newblock Mos: Towards scaling out-of-distribution detection for large semantic
  space.
\newblock In \emph{CVPR}, pages 8710--8719, 2021.

\bibitem[Huang et~al.(2021)Huang, Geng, and Li]{huang2021importance}
Rui Huang, Andrew Geng, and Yixuan Li.
\newblock On the importance of gradients for detecting distributional shifts in
  the wild.
\newblock \emph{NeurIPS}, 34, 2021.

\bibitem[Huang et~al.(2017)Huang, Li, Poursaeed, Hopcroft, and
  Belongie]{huang2017stacked}
Xun Huang, Yixuan Li, Omid Poursaeed, John Hopcroft, and Serge Belongie.
\newblock Stacked generative adversarial networks.
\newblock In \emph{CVPR}, pages 5077--5086, 2017.

\bibitem[Iandola et~al.(2017)Iandola, Han, Moskewicz, Ashraf, Dally, and
  Keutzer]{iandola2016squeezenet}
Forrest~N Iandola, Song Han, Matthew~W Moskewicz, Khalid Ashraf, William~J
  Dally, and Kurt Keutzer.
\newblock Squeezenet: Alexnet-level accuracy with 50x fewer parameters and< 0.5
  mb model size.
\newblock \emph{ICLR}, 2017.

\bibitem[Jiang et~al.(2022)Jiang, Sun, and Yu]{jiang2021revisiting}
Dihong Jiang, Sun Sun, and Yaoliang Yu.
\newblock Revisiting flow generative models for out-of-distribution detection.
\newblock In \emph{ICLR}, 2022.

\bibitem[Kim et~al.(2021)Kim, Shin, and Kim]{kim2021locally}
Keunseo Kim, JunCheol Shin, and Heeyoung Kim.
\newblock Locally most powerful bayesian test for out-of-distribution detection
  using deep generative models.
\newblock \emph{NeurIPS}, 34, 2021.

\bibitem[Kingma and Welling(2014)]{kingma2013auto}
Diederik~P Kingma and Max Welling.
\newblock Auto-encoding variational bayes.
\newblock \emph{ICLR}, 2014.

\bibitem[Kirichenko et~al.(2020)Kirichenko, Izmailov, and
  Wilson]{kirichenko2020normalizing}
Polina Kirichenko, Pavel Izmailov, and Andrew~G Wilson.
\newblock Why normalizing flows fail to detect out-of-distribution data.
\newblock \emph{NeurIPS}, 33:\penalty0 20578--20589, 2020.

\bibitem[Koh et~al.(2021)Koh, Sagawa, Marklund, Xie, Zhang, Balsubramani, Hu,
  Yasunaga, Phillips, Gao, et~al.]{koh2021wilds}
Pang~Wei Koh, Shiori Sagawa, Henrik Marklund, Sang~Michael Xie, Marvin Zhang,
  Akshay Balsubramani, Weihua Hu, Michihiro Yasunaga, Richard~Lanas Phillips,
  Irena Gao, et~al.
\newblock Wilds: A benchmark of in-the-wild distribution shifts.
\newblock In \emph{ICML}, pages 5637--5664. PMLR, 2021.

\bibitem[Kolesnikov et~al.(2020)Kolesnikov, Beyer, Zhai, Puigcerver, Yung,
  Gelly, and Houlsby]{kolesnikov2020big}
Alexander Kolesnikov, Lucas Beyer, Xiaohua Zhai, Joan Puigcerver, Jessica Yung,
  Sylvain Gelly, and Neil Houlsby.
\newblock Big transfer (bit): General visual representation learning.
\newblock In \emph{ECCV}, 2020.

\bibitem[Krizhevsky et~al.(2009)Krizhevsky, Hinton,
  et~al.]{krizhevsky2009learning}
Alex Krizhevsky, Geoffrey Hinton, et~al.
\newblock Learning multiple layers of features from tiny images.
\newblock 2009.

\bibitem[Kumar et~al.(2022)Kumar, Raghunathan, Jones, Ma, and
  Liang]{kumar2022fine}
Ananya Kumar, Aditi Raghunathan, Robbie Jones, Tengyu Ma, and Percy Liang.
\newblock Fine-tuning can distort pretrained features and underperform
  out-of-distribution.
\newblock \emph{ICLR}, 2022.

\bibitem[Lee et~al.(2018)Lee, Lee, Lee, and Shin]{lee2018simple}
Kimin Lee, Kibok Lee, Honglak Lee, and Jinwoo Shin.
\newblock A simple unified framework for detecting out-of-distribution samples
  and adversarial attacks.
\newblock \emph{NeurIPS}, 31, 2018.

\bibitem[Liang et~al.(2018)Liang, Li, and Srikant]{liang2017enhancing}
Shiyu Liang, Yixuan Li, and Rayadurgam Srikant.
\newblock Enhancing the reliability of out-of-distribution image detection in
  neural networks.
\newblock \emph{ICLR}, 2018.

\bibitem[Liu et~al.(2020)Liu, Wang, Owens, and Li]{liu2020energy}
Weitang Liu, Xiaoyun Wang, John Owens, and Yixuan Li.
\newblock Energy-based out-of-distribution detection.
\newblock \emph{NeurIPS}, 33:\penalty0 21464--21475, 2020.

\bibitem[Mar{\v{c}}enko and Pastur(1967)]{marvcenko1967distribution}
Vladimir~A Mar{\v{c}}enko and Leonid~Andreevich Pastur.
\newblock Distribution of eigenvalues for some sets of random matrices.
\newblock \emph{Mathematics of the USSR-Sbornik}, 1\penalty0 (4):\penalty0 457,
  1967.

\bibitem[Nalisnick et~al.(2019)Nalisnick, Matsukawa, Teh, Gorur, and
  Lakshminarayanan]{nalisnick2018deep}
Eric Nalisnick, Akihiro Matsukawa, Yee~Whye Teh, Dilan Gorur, and Balaji
  Lakshminarayanan.
\newblock Do deep generative models know what they don't know?
\newblock In \emph{ICLR}, 2019.

\bibitem[Nguyen et~al.(2015)Nguyen, Yosinski, and Clune]{nguyen2015deep}
Anh Nguyen, Jason Yosinski, and Jeff Clune.
\newblock Deep neural networks are easily fooled: High confidence predictions
  for unrecognizable images.
\newblock In \emph{CVPR}, pages 427--436, 2015.

\bibitem[Ovadia et~al.(2019)Ovadia, Fertig, Ren, Nado, Sculley, Nowozin,
  Dillon, Lakshminarayanan, and Snoek]{ovadia2019can}
Yaniv Ovadia, Emily Fertig, Jie Ren, Zachary Nado, David Sculley, Sebastian
  Nowozin, Joshua Dillon, Balaji Lakshminarayanan, and Jasper Snoek.
\newblock Can you trust your model's uncertainty? evaluating predictive
  uncertainty under dataset shift.
\newblock \emph{NeurIPS}, 32, 2019.

\bibitem[Qui{\~n}onero-Candela et~al.(2008)Qui{\~n}onero-Candela, Sugiyama,
  Schwaighofer, and Lawrence]{quinonero2008dataset}
Joaquin Qui{\~n}onero-Candela, Masashi Sugiyama, Anton Schwaighofer, and Neil~D
  Lawrence.
\newblock \emph{Dataset shift in machine learning}.
\newblock Mit Press, 2008.

\bibitem[Ren et~al.(2019)Ren, Liu, Fertig, Snoek, Poplin, Depristo, Dillon, and
  Lakshminarayanan]{ren2019likelihood}
Jie Ren, Peter~J Liu, Emily Fertig, Jasper Snoek, Ryan Poplin, Mark Depristo,
  Joshua Dillon, and Balaji Lakshminarayanan.
\newblock Likelihood ratios for out-of-distribution detection.
\newblock \emph{NeurIPS}, 32, 2019.

\bibitem[Rezende et~al.(2014)Rezende, Mohamed, and
  Wierstra]{rezende2014stochastic}
Danilo~Jimenez Rezende, Shakir Mohamed, and Daan Wierstra.
\newblock Stochastic backpropagation and approximate inference in deep
  generative models.
\newblock In \emph{ICML}, pages 1278--1286. PMLR, 2014.

\bibitem[Scheirer et~al.(2012)Scheirer, de~Rezende~Rocha, Sapkota, and
  Boult]{scheirer2012toward}
Walter~J Scheirer, Anderson de~Rezende~Rocha, Archana Sapkota, and Terrance~E
  Boult.
\newblock Toward open set recognition.
\newblock \emph{IEEE TPAMI}, 2012.

\bibitem[Schirrmeister et~al.(2020)Schirrmeister, Zhou, Ball, and
  Zhang]{schirrmeister2020understanding}
Robin Schirrmeister, Yuxuan Zhou, Tonio Ball, and Dan Zhang.
\newblock Understanding anomaly detection with deep invertible networks through
  hierarchies of distributions and features.
\newblock \emph{NeurIPS}, 33:\penalty0 21038--21049, 2020.

\bibitem[Sengupta and Mitra(1999)]{sengupta1999distributions}
Anirvan~M Sengupta and Partha~P Mitra.
\newblock Distributions of singular values for some random matrices.
\newblock \emph{Physical Review E}, 60\penalty0 (3):\penalty0 3389, 1999.

\bibitem[Serr{\`a} et~al.(2019)Serr{\`a}, {\'A}lvarez, G{\'o}mez, Slizovskaia,
  N{\'u}{\~n}ez, and Luque]{serra2019input}
Joan Serr{\`a}, David {\'A}lvarez, Vicen{\c{c}} G{\'o}mez, Olga Slizovskaia,
  Jos{\'e}~F N{\'u}{\~n}ez, and Jordi Luque.
\newblock Input complexity and out-of-distribution detection with
  likelihood-based generative models.
\newblock In \emph{ICLR}, 2019.

\bibitem[Sun et~al.(2021)Sun, Guo, and Li]{sun2021react}
Yiyou Sun, Chuan Guo, and Yixuan Li.
\newblock React: Out-of-distribution detection with rectified activations.
\newblock \emph{NeurIPS}, 34, 2021.

\bibitem[Sun et~al.(2020)Sun, Wang, Liu, Miller, Efros, and Hardt]{sun2020test}
Yu~Sun, Xiaolong Wang, Zhuang Liu, John Miller, Alexei Efros, and Moritz Hardt.
\newblock Test-time training with self-supervision for generalization under
  distribution shifts.
\newblock In \emph{ICML}, pages 9229--9248. PMLR, 2020.

\bibitem[Sutter et~al.(2021)Sutter, Krause, and Kuhn]{sutter2021robust}
Tobias Sutter, Andreas Krause, and Daniel Kuhn.
\newblock Robust generalization despite distribution shift via minimum
  discriminating information.
\newblock \emph{NeurIPS}, 34, 2021.

\bibitem[Tabak and Turner(2013)]{tabak2013family}
Esteban~G Tabak and Cristina~V Turner.
\newblock A family of nonparametric density estimation algorithms.
\newblock \emph{Communications on Pure and Applied Mathematics}, 66\penalty0
  (2):\penalty0 145--164, 2013.

\bibitem[Tack et~al.(2020)Tack, Mo, Jeong, and Shin]{tack2020csi}
Jihoon Tack, Sangwoo Mo, Jongheon Jeong, and Jinwoo Shin.
\newblock Csi: Novelty detection via contrastive learning on distributionally
  shifted instances.
\newblock \emph{NeurIPS}, 33:\penalty0 11839--11852, 2020.

\bibitem[Van~den Oord et~al.(2016)Van~den Oord, Kalchbrenner, Espeholt,
  Vinyals, Graves, et~al.]{van2016conditional}
Aaron Van~den Oord, Nal Kalchbrenner, Lasse Espeholt, Oriol Vinyals, Alex
  Graves, et~al.
\newblock Conditional image generation with pixelcnn decoders.
\newblock \emph{NeurIPS}, 29, 2016.

\bibitem[Van~Horn et~al.(2018)Van~Horn, Mac~Aodha, Song, Cui, Sun, Shepard,
  Adam, Perona, and Belongie]{van2018inaturalist}
Grant Van~Horn, Oisin Mac~Aodha, Yang Song, Yin Cui, Chen Sun, Alex Shepard,
  Hartwig Adam, Pietro Perona, and Serge Belongie.
\newblock The inaturalist species classification and detection dataset.
\newblock In \emph{CVPR}, 2018.

\bibitem[Vaze et~al.(2022)Vaze, Han, Vedaldi, and Zisserman]{vaze2021open}
Sagar Vaze, Kai Han, Andrea Vedaldi, and Andrew Zisserman.
\newblock Open-set recognition: A good closed-set classifier is all you need.
\newblock In \emph{ICLR}, 2022.

\bibitem[Wang et~al.(2021)Wang, Liu, Bocchieri, and Li]{wang2021can}
Haoran Wang, Weitang Liu, Alex Bocchieri, and Yixuan Li.
\newblock Can multi-label classification networks know what they don’t know?
\newblock \emph{NeurIPS}, 34, 2021.

\bibitem[Wang et~al.(2020)Wang, Dai, Wipf, and Zhu]{wang2020further}
Ziyu Wang, Bin Dai, David Wipf, and Jun Zhu.
\newblock Further analysis of outlier detection with deep generative models.
\newblock \emph{NeurIPS}, 33:\penalty0 8982--8992, 2020.

\bibitem[Wiles et~al.(2022)Wiles, Gowal, Stimberg, Alvise-Rebuffi, Ktena,
  Cemgil, et~al.]{wiles2021fine}
Olivia Wiles, Sven Gowal, Florian Stimberg, Sylvestre Alvise-Rebuffi, Ira
  Ktena, Taylan Cemgil, et~al.
\newblock A fine-grained analysis on distribution shift.
\newblock \emph{ICLR}, 2022.

\bibitem[Xiao et~al.(2010)Xiao, Hays, Ehinger, Oliva, and
  Torralba]{xiao2010sun}
Jianxiong Xiao, James Hays, Krista~A Ehinger, Aude Oliva, and Antonio Torralba.
\newblock Sun database: Large-scale scene recognition from abbey to zoo.
\newblock In \emph{CVPR}, 2010.

\bibitem[Xiao et~al.(2020)Xiao, Yan, and Amit]{xiao2020likelihood}
Zhisheng Xiao, Qing Yan, and Yali Amit.
\newblock Likelihood regret: An out-of-distribution detection score for
  variational auto-encoder.
\newblock \emph{NeurIPS}, 33:\penalty0 20685--20696, 2020.

\bibitem[Yang et~al.(2021)Yang, Zhou, Li, and Liu]{yang2021generalized}
Jingkang Yang, Kaiyang Zhou, Yixuan Li, and Ziwei Liu.
\newblock Generalized out-of-distribution detection: A survey.
\newblock \emph{arXiv preprint arXiv:2110.11334}, 2021.

\bibitem[Ye et~al.(2021)Ye, Xie, Cai, Li, Li, and Wang]{ye2021towards}
Haotian Ye, Chuanlong Xie, Tianle Cai, Ruichen Li, Zhenguo Li, and Liwei Wang.
\newblock Towards a theoretical framework of out-of-distribution
  generalization.
\newblock \emph{NeurIPS}, 34, 2021.

\bibitem[Yuan et~al.(2021)Yuan, Chen, Wang, Yu, Shi, Jiang, Tay, Feng, and
  Yan]{yuan2021tokens}
Li~Yuan, Yunpeng Chen, Tao Wang, Weihao Yu, Yujun Shi, Zi-Hang Jiang,
  Francis~EH Tay, Jiashi Feng, and Shuicheng Yan.
\newblock Tokens-to-token vit: Training vision transformers from scratch on
  imagenet.
\newblock In \emph{ICCV}, 2021.

\bibitem[Zaeemzadeh et~al.(2021)Zaeemzadeh, Bisagno, Sambugaro, Conci,
  Rahnavard, and Shah]{zaeemzadeh2021out}
Alireza Zaeemzadeh, Niccol{\`o} Bisagno, Zeno Sambugaro, Nicola Conci, Nazanin
  Rahnavard, and Mubarak Shah.
\newblock Out-of-distribution detection using union of 1-dimensional subspaces.
\newblock In \emph{CVPR}, pages 9452--9461, 2021.

\bibitem[Zhong et~al.(2021)Zhong, Fini, Roy, Luo, Ricci, and
  Sebe]{zhong2021neighborhood}
Zhun Zhong, Enrico Fini, Subhankar Roy, Zhiming Luo, Elisa Ricci, and Nicu
  Sebe.
\newblock Neighborhood contrastive learning for novel class discovery.
\newblock In \emph{CVPR}, pages 10867--10875, 2021.

\bibitem[Zhou et~al.(2017)Zhou, Lapedriza, Khosla, Oliva, and
  Torralba]{zhou2017places}
Bolei Zhou, Agata Lapedriza, Aditya Khosla, Aude Oliva, and Antonio Torralba.
\newblock Places: A 10 million image database for scene recognition.
\newblock \emph{IEEE TPAMI}, 2017.

\end{thebibliography}
}

\newpage
\appendix


\section{Experimental Setup}
\noindent \textbf{Implementation Details.} At the inference stage, all the images are resized to $480{\times}480$ for ResNetv2-101~\cite{he2016identity} and SqueezeNet~\cite{iandola2016squeezenet}. The source codes are implemented with \texttt{Pytorch 1.10.1}, and all experiments are run on a single NVIDIA Quadro RTX 6000 GPU. 

\noindent \textbf{Evaluation Metrics.} Following~\cite{huang2021mos,sun2021react,huang2021importance}, we measure the performance using two main metrics: (1) the false positive rate (FPR95) of OOD examples when the true positive rate of ID samples is at 95\%; and (2) the area under the receiver operating characteristic curve (AUROC). 

\begin{figure}[htbp]
    \centering
\begin{lstlisting}[language=Python]
#Our RankFeat (SVD) is applied on each individual \\
#feature matrix within the mini-batch.
feat = model.features(inputs)
B, C, H, W = feat.size()
feat = feat.view(B, C, H * W)
u,s,vt = torch.linalg.svd(feat)
feat = feat - s[:,0:1].unsqueeze(2)*u[:,:,0:1].bmm(vt[:,0:1,:])
feat = feat.view(B,C,H,W)
logits = model.classifier(feat)
score = torch.logsumexp(logits, dim=1)
\end{lstlisting}
    \caption{Pytorch-like codes of our \texttt{RankFeat} implementation.}
    \label{fig:code}
\end{figure}

\noindent \textbf{Pseudo Code of RankFeat.} Fig.~\ref{fig:code} presents the Pytorch-like implementation of our \texttt{RankFeat}. We use \texttt{torch.linalg.svd} to conduct SVD on each individual feature matrix in the mini-batch.
\section{More Evaluation Results}




\subsection{Large-scale Species Dataset}

\begin{table}[htbp]
    \caption{The evaluation results on four sub-sets of Species~\cite{hendrycks2019scaling} based on ResNetv2-101~\cite{he2016identity}. All values are reported in percentages, and these \emph{post hoc} methods are directly applied to the model pre-trained on ImageNet-1k~\cite{deng2009imagenet}. The best three results are highlighted with \textbf{\textcolor{red}{red}}, \textbf{\textcolor{blue}{blue}}, and \textbf{\textcolor{cyan}{cyan}}.}
    \centering
    \resizebox{0.99\linewidth}{!}{
    \begin{tabular}{c|cc|cc|cc|cc|cc}
    \toprule
        \multirow{3}*{\textbf{Methods}} & \multicolumn{2}{c|}{\textbf{Protozoa}} & \multicolumn{2}{c|}{\textbf{Microorganisms}} & \multicolumn{2}{c|}{\textbf{Plants}} & \multicolumn{2}{c|}{\textbf{Mollusks}} & \multicolumn{2}{c}{\textbf{Average}}   \\
        \cmidrule{2-11}
         & FPR95 & AUROC  & FPR95 & AUROC  & FPR95 & AUROC  & FPR95 & AUROC & FPR95 & AUROC  \\
         & ($\downarrow$) & ($\uparrow$) & ($\downarrow$) & ($\uparrow$) & ($\downarrow$) & ($\uparrow$) & ($\downarrow$) & ($\uparrow$) & ($\downarrow$) & ($\uparrow$)\\
    \midrule
    MSP~\cite{hendrycks2016baseline} &75.81&83.20&72.23&84.25& 61.48 & 87.78 &85.62&70.51 & 73.79 & 81.44 \\
    ODIN~\cite{liang2017enhancing} &75.97&\textbf{\textcolor{cyan}{85.11}}&65.94&89.35&55.69&90.79  &86.22&71.31 & 70.96&84.14\\
    Energy~\cite{liu2020energy} &79.49&84.34 &60.87&\textbf{\textcolor{blue}{90.30}} &54.67&90.95&88.47&70.53&70.88&84.03 \\
    ReAct~\cite{sun2021react} &81.74&84.26&58.82&85.88&\textbf{\textcolor{red}{36.90}}&\textbf{\textcolor{red}{93.78}}&90.58&\textbf{\textcolor{cyan}{76.33}}&67.02&\textbf{\textcolor{cyan}{85.06}}\\
    \midrule
    \rowcolor{gray!20}\textbf{RankFeat (Block 4)} & \textbf{\textcolor{cyan}{66.98}}&70.19&\textbf{\textcolor{blue}{39.06}}&86.67&\textbf{\textcolor{cyan}{46.31}}&79.98&\textbf{\textcolor{blue}{80.14}}&59.92&\textbf{\textcolor{blue}{58.12}}&74.19\\
    \rowcolor{gray!20}\textbf{RankFeat (Block 3)} & \textbf{\textcolor{blue}{58.99}}&\textbf{\textcolor{red}{88.81}}&\textbf{\textcolor{cyan}{49.72}}&\textbf{\textcolor{cyan}{90.04}}&47.01&\textbf{\textcolor{cyan}{91.85}}&\textbf{\textcolor{cyan}{80.37}}&\textbf{\textcolor{red}{79.61}}&\textbf{\textcolor{cyan}{59.02}}&\textbf{\textcolor{blue}{87.58}}\\
     \rowcolor{gray!20} \textbf{RankFeat (Block 3 + 4)}& \textbf{\textcolor{red}{52.78}}&\textbf{\textcolor{blue}{88.65}}&\textbf{\textcolor{red}{37.21}}&\textbf{\textcolor{red}{92.82}}&\textbf{\textcolor{blue}{38.07}}&\textbf{\textcolor{blue}{92.88}}&\textbf{\textcolor{red}{76.38}}&\textbf{\textcolor{blue}{78.13}}&\textbf{\textcolor{red}{51.11}}&\textbf{\textcolor{red}{88.37}}\\
    \bottomrule
    \end{tabular}
    }
    \label{tab:species_res101}
\end{table}

The Species~\cite{hendrycks2019scaling} dataset is a large-scale OOD validation benchmark consisting of $71,3449$ images, which is designed for ImageNet-1k~\cite{deng2009imagenet} and ImageNet 21-k~\cite{kolesnikov2020big} as the ID sets. We select four sub-sets as the OOD benchmark, namely \texttt{Protozoa}, \texttt{Microorganisms}, \texttt{Plants}, and \texttt{Mollusks}. Table~\ref{tab:species_res101} present the evaluation results. Our \texttt{RankFeat} achieves the best performance, surpassing other methods by $\textbf{15.91\%}$ in the average FPR95 and by $\textbf{3.31\%}$ in the average AUROC.

\subsection{CIFAR100 with Different Architectures}

\begin{table}[htbp]
    \centering 
    \caption{The evaluation results with different model architectures on CIFAR100~\cite{krizhevsky2009learning}. All values are reported in percentages, and these \emph{post hoc} methods are directly applied to the model. The best two results are highlighted with \textbf{\textcolor{red}{red}} and \textbf{\textcolor{blue}{blue}}.}
    \resizebox{0.99\linewidth}{!}{
    \begin{tabular}{c|c|cc|cc|cc|cc|cc}
    \toprule
        \multirow{3}*{\textbf{Model}}&\multirow{3}*{\textbf{Methods}} & \multicolumn{2}{c|}{\textbf{iNaturalist}} & \multicolumn{2}{c|}{\textbf{SUN}} & \multicolumn{2}{c|}{\textbf{Places}} & \multicolumn{2}{c|}{\textbf{Textures}} & \multicolumn{2}{c}{\textbf{Average}}   \\
        \cmidrule{3-12}
         && FPR95 & AUROC  & FPR95 & AUROC  & FPR95 & AUROC  & FPR95 & AUROC & FPR95 & AUROC  \\
         && ($\downarrow$) & ($\uparrow$) & ($\downarrow$) & ($\uparrow$) & ($\downarrow$) & ($\uparrow$) & ($\downarrow$) & ($\uparrow$) & ($\downarrow$) & ($\uparrow$)\\
    \midrule
    \multirow{7}*{\textbf{RepVGG-A0~\cite{ding2021repvgg}}}&MSP~\cite{hendrycks2016baseline} 
    &61.55&85.03&91.05&69.19&65.45&\textbf{\textcolor{blue}{82.10}}&86.68&65.56&76.18&75.47\\
    &ODIN~\cite{liang2017enhancing} 
    &50.20&87.88&88.00&66.56&61.85&79.34&84.87&63.89&71.23&74.42\\
    &Energy~\cite{liu2020energy} 
    &53.71&84.59&86.71&66.58&59.71&78.64&84.57&63.88&71.18&73.42\\
    &Mahalanobis~\cite{lee2018simple} 
    &81.43&74.81&89.77&67.12&79.49&73.06&\textbf{\textcolor{blue}{64.95}}&\textbf{\textcolor{blue}{82.19}}&78.91&74.30\\
    &GradNorm~\cite{huang2021importance} 
    &78.87&68.21&95.10&44.73&66.25&75.41&92.98&43.83&83.30&58.05\\
    &ReAct~\cite{sun2021react} 
    &\textbf{\textcolor{blue}{48.09}}&\textbf{\textcolor{red}{93.00}}&\textbf{\textcolor{blue}{73.87}}&\textbf{\textcolor{red}{78.12}}&\textbf{\textcolor{blue}{61.63}}&78.43&75.23&81.36&\textbf{\textcolor{blue}{64.71}}&\textbf{\textcolor{blue}{82.73}}\\
    \cmidrule{2-12}
     \rowcolor{gray!20} \cellcolor{white}&\textbf{RankFeat}&\textbf{\textcolor{red}{40.19}}&\textbf{\textcolor{blue}{88.06}}&\textbf{\textcolor{red}{70.47}}&\textbf{\textcolor{blue}{76.35}}&\textbf{\textcolor{red}{57.75}}&\textbf{\textcolor{red}{83.58}}&\textbf{\textcolor{red}{52.89}}&\textbf{\textcolor{red}{83.28}}&\textbf{\textcolor{red}{55.33}} &\textbf{\textcolor{red}{82.82}}\\
    
    \midrule
    \multirow{7}*{\textbf{ResNet-56~\cite{he2016deep}}}&MSP~\cite{hendrycks2016baseline} 
    &77.69&78.25&93.54&66.93&81.57&76.71&88.47&65.79&85.32&71.92\\
    &ODIN~\cite{liang2017enhancing} 
    &66.92&79.25&95.05&50.45&77.45&72.88&90.51&53.47&82.48&64.01\\
    &Energy~\cite{liu2020energy} 
    &65.24&79.13&95.05&49.33&77.10&72.32&90.39&52.68&81.95&63.37\\
    &Mahalanobis~\cite{lee2018simple} 
    &89.47&69.32&91.38&54.76&82.32&77.53&\textbf{\textcolor{blue}{68.83}}&\textbf{\textcolor{blue}{79.64}}&83.00&70.31\\
    &GradNorm~\cite{huang2021importance} 
    &96.72&42.09&94.19&47.97&94.61&48.09&89.14&50.18&93.67&47.08\\
    &ReAct~\cite{sun2021react} 
    &\textbf{\textcolor{blue}{50.59}}&\textbf{\textcolor{red}{90.56}}&\textbf{\textcolor{blue}{69.23}}&\textbf{\textcolor{red}{85.79}}&\textbf{\textcolor{blue}{55.38}}&\textbf{\textcolor{blue}{87.98}}&82.60&75.51&\textbf{\textcolor{blue}{64.50}}&\textbf{\textcolor{blue}{84.96}}\\
    \cmidrule{2-12}
    \rowcolor{gray!20} \cellcolor{white} &\textbf{RankFeat} &\textbf{\textcolor{red}{34.62}}&\textbf{\textcolor{blue}{88.21}}&\textbf{\textcolor{red}{61.82}}&\textbf{\textcolor{blue}{80.50}}&\textbf{\textcolor{red}{53.79}}&\textbf{\textcolor{red}{89.71}}&\textbf{\textcolor{red}{30.89}}&\textbf{\textcolor{red}{91.31}}&\textbf{\textcolor{red}{45.28}}&\textbf{\textcolor{red}{87.43}}\\
    \bottomrule
    \end{tabular}
    }
    \label{tab:results_cifar}
\end{table}

We also evaluate our method on the CIFAR benchmark with various model architectures. The evaluation OOD datasets are the same with those of the ImageNet-1k benchmark. We take ResNet-56~\cite{he2016deep} and RepVGG-A0~\cite{ding2021repvgg} pre-trained on ImageNet-1k as the backbones, and then fine-tune them on CIAR100~\cite{krizhevsky2009learning} for $100$ epochs. The learning rate is initialized with $0.1$ and is decayed by $10$ every $30$ epoch. Notice that this training process is to obtain a well-trained classifier but the ODO methods (including ours) are still \emph{post hoc} and do not need any extra training.

Table~\ref{tab:results_cifar} compares the performance against all the \emph{post hoc} baselines. Our \texttt{RankFeat} establishes the \textit{state-of-the-art} performances across architectures on most datasets and metrics, outperforming the second best method by \textbf{9.38 \%} in the average FPR95 on RepVGG-A0 and by \textbf{19.22 \%} in the average FPR95 on ResNet-56. Since the CIFAR images are small in resolution (\emph{i.e.,} $32{\times}32$), the downsampling times and the number of feature blocks of the original models are reduced. Hence we only apply \texttt{RankFeat} to the final feature before the last GAP layer.

\subsection{One-class CIFAR10}

To further demonstrate the applicability of our method, we follow~\cite{emmott2013systematic,golan2018deep,tack2020csi} and conduct experiments on one-class CIFAR10. The setup is as follows: we choose one
of the classes as the ID set while keeping other classes as OOD sets. Table~\ref{tab:oneclass_cifar10} reports the average AUROC on CIFAR10. Our \texttt{RankFeat} outperforms other baselines on most sub-set as well as on the average result.

\begin{table}[htbp]
    \centering
    \caption{The average AUROC (\%) on one-class CIFAR10 based on ResNet-56.}
    \resizebox{0.99\linewidth}{!}{
    \begin{tabular}{c|cccccccccc|c}
    \toprule
         Methods & Plane & Car & Bird & Cat & Deer & Dog & Frog & Horse & Ship & Truck & Mean \\
    \midrule
         MSP    &59.75 &52.48 &62.96 &48.73 &59.15 &52.39 &67.33 &59.34 &54.55 &51.97 &56.87 \\
         Energy &\textbf{83.12} &91.56 &68.99 &56.02 &75.03 &77.33 &69.50 &88.41 &82.88 &84.74 &77.76 \\
         ReAct  &82.24 &96.69 &78.32 &76.84 &76.11 &86.80 &86.15 &90.95 &89.91 &\textbf{94.17} &85.82 \\
         \rowcolor{gray!20}\textbf{RankFeat} &79.26 &\textbf{98.54} &\textbf{82.04} &\textbf{80.28} &\textbf{82.89} &\textbf{90.28} &\textbf{89.06} &\textbf{95.30} &\textbf{94.11} &94.02 &\textbf{88.58} \\
    \bottomrule
    \end{tabular}
    }
    \label{tab:oneclass_cifar10}
\end{table}





\section{Baseline Methods}

For the convenience of audiences, we briefly recap the previous \emph{post hoc} methods for OOD detection. Some implementation details of the methods are also discussed.

\noindent \textbf{MSP~\cite{hendrycks2016baseline}.} One of the earliest work considered directly using the Maximum Softmax Probability (MSP) as the scoring function for OOD detection. Let $f(\cdot)$ and $\mathbf{x}$ denote the model and input, respectively. The MSP score can be computed as:
\begin{equation}
    \texttt{MSP}(\mathbf{x}) = \max\Big({\rm Softmax}(f(\mathbf{x}))\Big)
\end{equation}
Despite the simplicity of this approach, the MSP score often fails as neural networks could assign arbitrarily high confidences to the OOD data~\cite{nguyen2015deep}.

\noindent \textbf{ODIN~\cite{liang2017enhancing}.} Based on MSP~\cite{hendrycks2016baseline}, ODIN~\cite{liang2017enhancing} further integrated temperature scaling and input perturbation to better separate the ID and OOD data. The ODIN score is calculated as:
\begin{equation}
    \texttt{ODIN}(\mathbf{x}) = \max\Big({\rm Softmax}(\frac{f(\mathbf{\bar{x}})}{T})\Big)
\end{equation}
where $T$ is the hyper-parameter temperature, and $\mathbf{\bar{x}}$ denote the perturbed input. Following the setting in~\cite{huang2021importance}, we set $T{=}1000$. According to~\cite{huang2021importance}, the input perturbation does not bring any performance improvement on the ImageNet-1k benchmark. Hence, we do not perturb the input either.

\noindent \textbf{Energy score~\cite{liu2020energy}.} Liu \emph{et al.}~\cite{liu2020energy} argued that an energy score is superior than the MSP because it is theoretically aligned with the input probability density, \emph{i.e.,} the sample with a higher energy correspond to data with a lower likelihood of occurrence. Formally, the energy score maps the logit output to a scalar function as:
\begin{equation}
    \texttt{Energy}(\mathbf{x}) = \log\sum^{C}_{i=1}\exp(f_{i}(\mathbf{x}))
\end{equation}
where $C$ denotes the number of classes.

\noindent \textbf{Mahalanobis distance~\cite{lee2018simple}.} Lee~\emph{et al.}~\cite{lee2018simple} proposed to model the Softmax outputs as the mixture of multivariate Gaussian distributions and use the Mahalanobis distance as the scoring function for OOD uncertainty estimation. The score is computed as:
\begin{equation}
    \texttt{Mahalanobis}(\mathbf{x}) = \max_{i}\Big(-(f(\mathbf{x})-\mu_{i})^{T}\Sigma(f(\mathbf{x})-\mu_{i})\Big)
\end{equation}
where $\mu_{i}$ denotes the feature vector mean, and $\Sigma$ represents the covariance matrix across classes. Following~\cite{huang2021importance}, we use $500$ samples randomly selected from ID datasets and an auxiliary tuning dataset to train the logistic regression and tune the perturbation strength $\epsilon$. For the tuning dataset, we use FGSM~\cite{goodfellow2014explaining} with a perturbation size of 0.05 to generate adversarial examples. The selected $\epsilon$ is set as $0.001$ for ImageNet-1k.

\noindent \textbf{GradNorm~\cite{huang2021importance}.} Huang~\emph{et al.}~\cite{huang2021importance} proposed to estimate the OOD uncertainty by utilizing information extracted from the gradient space. They compute the KL divergence between the {\rm Softmax} output and a uniform distribution, and back-propagate the gradient to the last layer. Then the vector norm of the gradient is used as the scoring function. Let $\mathbf{w}$ and $\mathbf{u}$ denote the weights of last layer and the uniform distribution. The score is calculated as:
\begin{equation}
    \texttt{GradNorm}(\mathbf{x}) = ||\frac{\partial D_{KL}(\mathbf{u}||{\rm Softmax}(f(\mathbf{x})))}{\partial \mathbf{w}}||_{1}
\end{equation}
where $||\cdot||_{1}$ denotes the $L_{1}$ norm, and $D_{KL}(\cdot)$ represents the KL divergence measure.

\noindent \textbf{ReAct~\cite{sun2021react}.} In~\cite{sun2021react}, the authors observed that the activations of the penultimate layer are quite different for ID and OOD data. The OOD data is biased towards triggering very high activations, while the ID data has the well-behaved mean and deviation. In light of this finding, they propose to clip the activations as:
\begin{equation}
    \max(f_{l{-}1}(\mathbf{x}),\tau)
\end{equation}
where $f_{l{-}1}(\cdot)$ denotes the activations for the penultimate layer, and $\tau$ is the upper limit computed as the $90$-th percentile of activations of the ID data. Finally, the Energy score~\cite{liu2020energy} is computed for estimating the OOD uncertainty.


\section{Visualization about RankFeat}

\begin{figure}[htbp]
    \centering
    \includegraphics[width=0.99\linewidth]{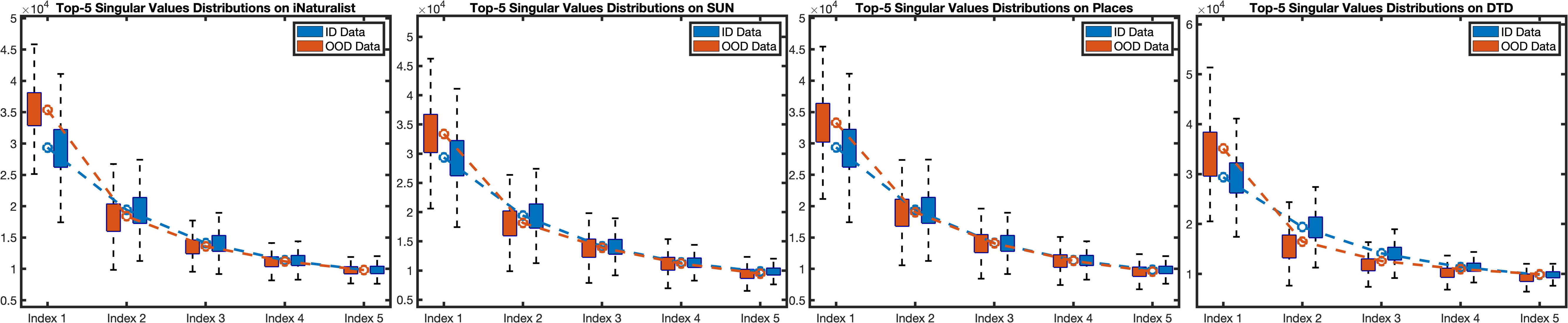}
    \caption{The top-5 singular value distribution of the ID dataset and OOD datasets. The first singular values $\mathbf{s}_{1}$ of OOD data are consistently much larger than those of ID data on each OOD dataset.}
    \label{fig:singu_dist}
\end{figure}

\subsection{Singular Value Distribution}

Fig.~\ref{fig:singu_dist} compares the top-5 singular value distribution of ID and OOD feature matrices on all the datasets. Our novel observation consistently holds for every OOD dataset: the dominant singular value $\mathbf{s}_{1}$ of OOD feature always tends to be significantly larger than that of ID feature.

\begin{figure}[htbp]
    \centering
    \includegraphics[width=0.9\linewidth]{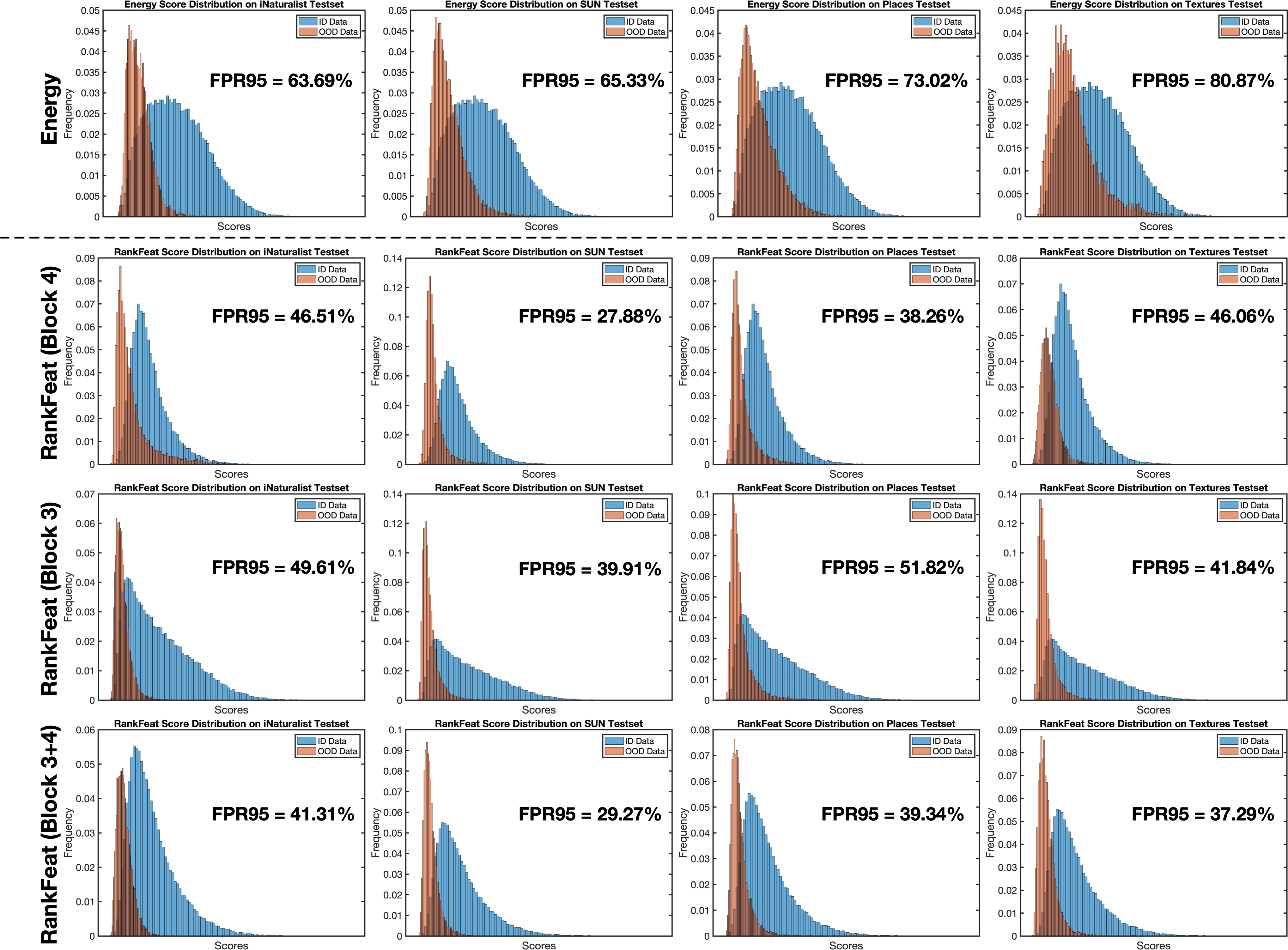}
    \caption{The score distributions of \texttt{Energy}~\cite{liu2020energy} (top row) and our proposed \texttt{RankFeat} (rest rows) on four OOD datasets. Our \texttt{RankFeat} applies to different high-level features at the later depths of the network, and their score functions can be further fused. }
    \label{fig:score_dist_complete}
\end{figure}

\subsection{Score Distribution}

Fig.~\ref{fig:score_dist_complete} displays the score distributions of \texttt{RankFeat} at Block 3 and  Block 4, as well as the fused results. Our \texttt{RankFeat} works for both high-level features. For the score fusion, when Block 3 and Block 4 features are of similar scores $(diff.{<}5\%)$, the feature combination could have further improvements.  

\subsection{Output Distribution}

Fig.~\ref{fig:output_logit}(a) presents the output distribution (\emph{i.e.,} the logits after \texttt{Softmax} layer) on \texttt{ImageNet} and \texttt{iNaturalist}. After our \texttt{RankFeat}, the OOD data have a larger reduction in the probability output; most of OOD predictions are of very small probabilities (${<}0.1$).

\subsection{Logit Distribution}

\begin{figure}[htbp]
    \centering
    \includegraphics[width=0.99\linewidth]{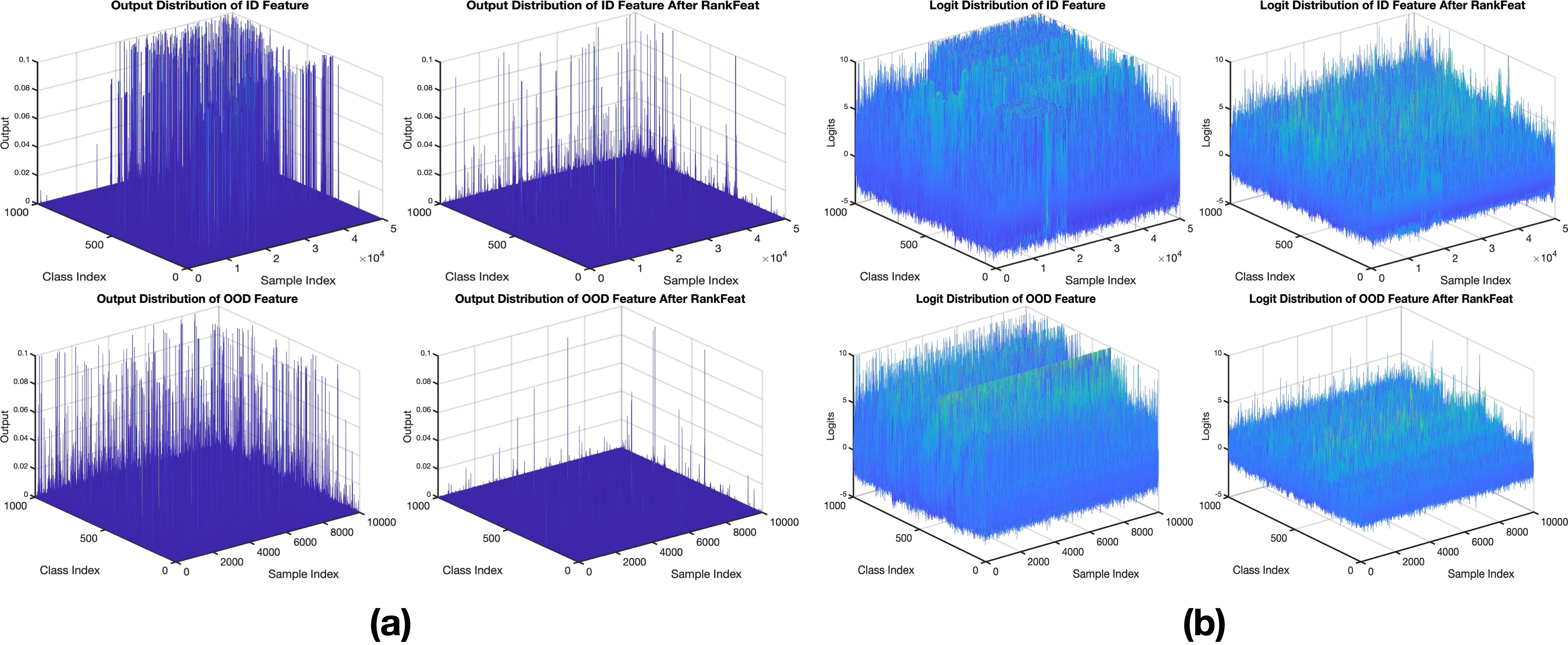}
    \caption{\textbf{(a)} Output distributions of \texttt{RankFeat}. \textbf{(b)} Logit distributions of \texttt{RankFeat}. }
    \label{fig:output_logit}
\end{figure}

Fig.~\ref{fig:output_logit}(b) displays the logits distribution of our \texttt{RankFeat}. The OOD logits after \texttt{RankFeat} have much less variations and therefore are closer to the uniform distribution.

\section{Why are the singular value distributions of ID and OOD features different?}

In the paper, we give some theoretical analysis to explain the working mechanism of our \texttt{RankFeat}. It would be also interesting to investigate why the singular value distributions of the ID and OOD features are different. Here we give an intuitive conjecture. Since the network is well trained on the ID training set, when encountered with ID data, the feature matrix is likely to be more informative. Accordingly, more singular vectors would be active and the matrix energies spread over the corresponding singular values, leading to a more flat spectrum. On the contrary, for the unseen OOD data, the feature is prone to have a more compact representation, and less singular vectors might be active. In this case, the dominant singular value of OOD feature would be larger and would take more energies of the matrix. The informativeness can also be understood by considering applying PCA on the feature matrix. Suppose that we are using PCA to reduce the dimension of ID and OOD feature to $1$. The amount of retained information can be measured by explained variance (\%). The metric is defined as $\sum_{i=0}^{k}\mathbf{s}_{i}^2/\sum_{j=0}^{n}\mathbf{s}_{j}^2$ where $k$ denotes the projected dimension and $n$ denotes the total dimension. It measures the portion of variance that the projected data could account for. We compute the average explained variance of all datasets and present the result in Table~\ref{tab:explained_varaince}.

\begin{table}[h]
    \centering
    \caption{The average explained variance ratio (\%) of the ID and OOD datasets.}
    \begin{tabular}{c|c|c|c|c|c}
    \toprule
         Dataset & ImageNet-1k & iNaturalist & SUN & Places & Textures  \\
    \midrule
         Explained Variance (\%) & \textbf{28.57} & 38.74 & 35.79 & 35.17 & 42.21 \\
    \bottomrule
    \end{tabular}
    \label{tab:explained_varaince}
\end{table}

As can be observed, the OOD datasets have a larger explained variance ratio than the ID dataset. \emph{That being said, to retain the same amount of information, we need fewer dimensions for the projection of OOD features. This indicates that the information of OOD feature is easier to be captured and the OOD feature matrix is thus less informative.}

\textit{As for how the training leads to the difference, we doubt that the well-trained network weights might cause and amplify the gap in the dominant singular value of the ID and OOD feature.} To verify this guess, we compute the singular values distributions 
of the Google BiT-S ResNetv2-100 model~\cite{he2016identity,kolesnikov2020big} with different training steps, as well as a randomly initialized network as the baseline. 

\begin{figure}[htbp]
    \centering
    \includegraphics[width=0.99\linewidth]{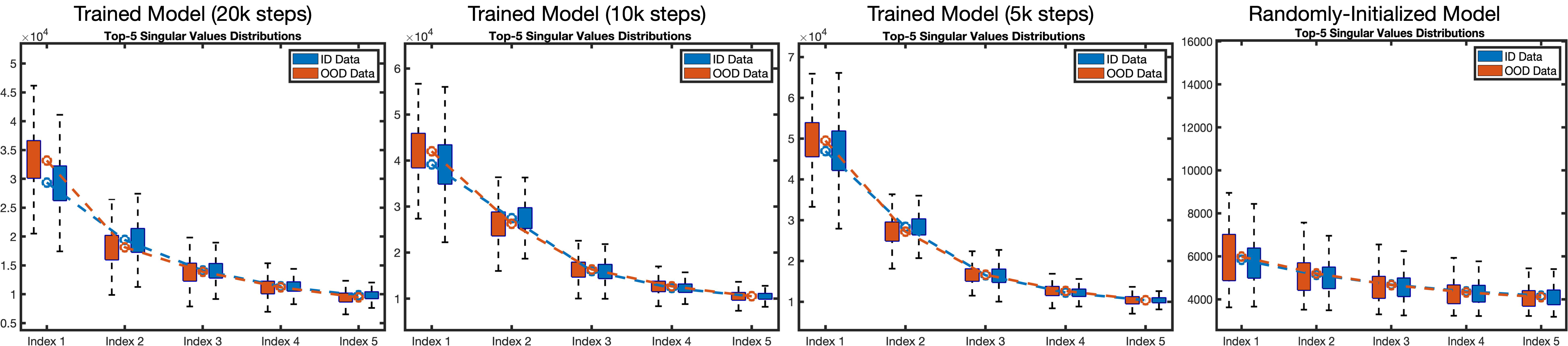}
    \caption{The top-5 largest singular value distributions of the pre-trained network with different training steps. For the untrained network initialized with random weights, the singular values distributions of ID and OOD feature exhibit very similar behaviors. As the training step increases, the difference between the largest singular value is gradually amplified.}
    \label{fig:rand_train}
\end{figure}

Fig.~\ref{fig:rand_train} depicts the top-5 largest singular value distributions of the network with different training steps. Unlike the trained networks, the untrained network with random weights has quite a similar singular value distribution for the ID and OOD data. The singular values of both ID and OOD features are of similar magnitudes with the untrained network. However, when the number of training steps is increased, the gap of dominant singular value between ID and OOD feature is magnified accordingly.
This phenomenon supports our conjecture that the well-trained network weights cause and amplify the difference of the largest singular value. Interestingly, our finding is coherent with~\cite{vaze2021open}. In~\cite{vaze2021open}, the authors demonstrate that the classification accuracy of a model is highly correlated with its ability of OOD detection and open-set recognition. Training a stronger model could naturally improve the OOD detection performance. We empirically show that the gap of the dominant singular value is gradually amplifying as the training goes on, which serves as supporting evidence for~\cite{vaze2021open}.

\section{Theorem and Proof of Manchenko-Pastur Law}

In the paper, we use the MP distribution of random matrices to show that removing the rank-1 matrix makes the statistics of OOD features closer to random matrices. For self-containment and readers' convenience, here we give a brief proof of Manchenko-Pastur Law.

\begin{thm}
Let $\mathbf{X}$ be a random matrix of shape $t{\times}n$ whose entries are random variables with $E(\mathbf{X}_{ij}=0)$ and $E(\mathbf{X}_{ij}^2=1)$. Then the eigenvalues of the sample covariance $\mathbf{Y}=\frac{1}{n}\mathbf{X}\mathbf{X}^{T}$ converges to the probability density function: $\rho(\lambda) = \frac{t}{n} \frac{\sqrt{(\lambda_{+}-\lambda)(\lambda-\lambda_{-})}}{2\pi\lambda\sigma^2}\ for\  \lambda\in[\lambda_{-},\lambda_{+}]$ where $\lambda_{-}{=}\sigma^{2} (1-\sqrt{\frac{n}{t}})^2$ and $ \lambda_{+}{=}\sigma^{2} (1+\sqrt{\frac{n}{t}})^2$.
\end{thm}
\begin{proof}

Similar with the deduction of our bound analysis, the sample covariance $\mathbf{Y}$ can be written as the sum of rank-1 matrices:
\begin{equation}
    \mathbf{Y}=\sum_{s=0}^{t}=\mathbf{Y}^{s}_{n},\ \mathbf{Y}^{s}_{n}=\mathbf{U}^{s}_{n}\mathbf{D}^{s}_{n}(\mathbf{U}^{s}_{n})^{*}
\end{equation}
where $\mathbf{U}^{s}_{n}$ is a unitary matrix, and $\mathbf{D}^{s}_{n}$ is a diagonal matrix with the only eigenvalue $\beta=\nicefrac{n}{t}$ for large $n$ (rank-1 matrix). Then we can compute the Stieltjes transform of each $\mathbf{Y}^{s}_{n}$ as:
\begin{equation}
    s_{n}(z)=\frac{1}{n}{\rm tr}(\mathbf{Y}^{s}_{n}-z\mathbf{I})^{-1}
\end{equation}
Relying on Neumann series, the above equation can be re-written as:
\begin{equation}
    s_{n}(z)=-\frac{1}{n}\sum_{k=0}^{\infty}\frac{{\rm tr}(\mathbf{Y}^{s}_{n})^t}{z^{k+1}}=-\frac{1}{n}\Big(\frac{n}{z}+\sum_{k=1}^{\infty}\frac{\beta^{k}}{z^{k+1}}\Big)=-\frac{1}{n}\Big(\frac{n-1}{z}+\frac{1}{z-\beta}\Big)
\end{equation}
Let $z:=z_{n}(s)$ and we can find the function inverse of the transform:
\begin{equation}
    n s z_{n}(s)^2-n(s\beta-1)z_{n}(s)-(n-1)\beta=0
\end{equation}
The close-formed solution is calculated as:
\begin{equation}
\begin{aligned}
    z_{n}(s) &= \frac{n(s\beta-1)\pm\sqrt{n^2(s\beta-1)^2 + 4n(n-1)s\beta}}{2ns}\\
             &\approx \frac{1}{2ns}\Big(n(s\beta-1)\pm\Big|n(s\beta+1)-\cancel{\frac{2s\beta}{\beta+1}}\Big|\Big)
\end{aligned}
\end{equation}
For large $n$, the term $\frac{2s\beta}{\beta+1}$ is sufficiently small and we can omit it. The solution is defined as:
\begin{equation}
    z_{n}(s) = - \frac{1}{s} + \frac{\beta}{n(1+s\beta)}
\end{equation}
The R transform of each $\mathbf{Y}^{s}_{n}$ is given by:
\begin{equation}
    R_{\mathbf{Y}^{s}_{n}}(s) = z_{n}(-s)-\frac{1}{s}=\frac{\beta}{n(1-s\beta)} 
\end{equation}
Accordingly, the R transform for $\mathbf{Y}_{n}$ is given by:
\begin{equation}
    R_{\mathbf{Y}}(s) = t R_{\mathbf{Y}^{s}_{n}}(s) = \frac{\beta t}{n(1-s\beta)}=\frac{1}{1-s\beta}
\end{equation}
Thus, the inverse Stieltjes transform of $\mathbf{Y}$ is
\begin{equation}
    z(s)=-\frac{1}{s} + \frac{1}{1+s\beta}
\end{equation}
Then the Stieltjes transform of $\mathbf{Y}$ is computed by inverting the above equation as:
\begin{equation}
    s(z)=\frac{-(z+\beta+1)+\sqrt{(z+\beta+1)^2-4\beta z}}{2z\beta}
\end{equation}
Since $\beta=\nicefrac{b}{t}$, finding the limiting distribution of the above equation directly gives the Manchenko-Pastur distribution:
\begin{equation}
\begin{gathered}
    \rho(\lambda) = \frac{t}{n} \frac{\sqrt{(\lambda_{+}-\lambda)(\lambda-\lambda_{-})}}{2\pi\lambda\sigma^2}\ for\  \lambda\in[\lambda_{-},\lambda_{+}], \\
    \lambda_{-}{=}\sigma^{2} (1-\sqrt{\frac{n}{t}})^2, \lambda_{+}{=}\sigma^{2} (1+\sqrt{\frac{n}{t}})^2
\end{gathered}
\end{equation}
The theorem is thus proved.
\end{proof}



\end{document}